\documentclass{article}

\usepackage{microtype}
\usepackage{graphicx}
\usepackage{subfigure}
\usepackage{booktabs} 
\usepackage{colortbl}

\usepackage{fullpage}
\usepackage{breakcites}
\usepackage{natbib}
\usepackage{algorithm}
\usepackage{algorithmic}

\usepackage{hyperref}
\hypersetup{
  colorlinks   = true, 
  urlcolor     = blue, 
  linkcolor    = blue, 
  citecolor    = blue   
}

\newcommand{\prraman}[1]{{\color[rgb]{0,0,1}[prraman: #1]}}

\usepackage{amsmath}
\usepackage{amssymb}
\usepackage{mathtools}
\usepackage{amsthm}
\usepackage{bm}
\usepackage{enumitem}

\usepackage[capitalize,noabbrev]{cleveref}

\theoremstyle{plain}
\newtheorem{theorem}{Theorem}[section]

\newtheorem{lemma}[theorem]{Lemma}
\newtheorem{corollary}[theorem]{Corollary}
\theoremstyle{definition}

\newtheorem{assumption}[theorem]{Assumption}
\theoremstyle{remark}

\def\dpos{{\cal D}_{\rm pos}}
\def\dneg{{\bf D}_{\rm neg}}
\def\dx{{\cal D}_{X}}

\def\px{{\cal P}_{x,\theta}}
\def\data{{\bf D}}
\def\mneg{m_{\rm neg}}
\def\smath#1{\text{\scalebox{.8}{$#1$}}}
\def\algname{EMC$^2$}
\newcommand{\tworows}[2]{\begin{tabular}{@{}l@{}} {{#1}} \\ {#2} \end{tabular}}
\def\ker{{\bf K}}
\newcommand*\samethanks[1][\value{footnote}]{\footnotemark[#1]}

\def\papertitle{\algname: Efficient MCMC Negative Sampling for Contrastive Learning with Global Convergence}

\title{\papertitle}
\author{Chung-Yiu Yau\thanks{C.-Y.~Yau and H.-T.~Wai are with the Department of Systems Engineering and Engineering Management, The Chinese University of Hong Kong, Hong Kong SAR of China.} \and Hoi-To Wai\samethanks \and Parameswaran Raman\thanks{P.~Raman and S.~Sarkar are with Amazon Web Services, USA.} \and Soumajyoti Sarkar\samethanks \and Mingyi Hong\thanks{M.~Hong is with Department of Electrical and Computer Engineering, University of Minnesota, USA.}}

\usepackage[textsize=tiny]{todonotes}

\begin{document}

\maketitle
\renewcommand*{\thefootnote}{\fnsymbol{footnote}}
\setcounter{footnote}{4}
\footnotetext{The work of C.-Y. Yau was done while interning at Amazon Web Services. M. Hong holds concurrent appointments as an Amazon Scholar and as a faculty at University of Minnesota. This paper describes their work performed at Amazon.}
\renewcommand*{\thefootnote}{\arabic{footnote}}
\setcounter{footnote}{0}

\begin{abstract}  
A key challenge in contrastive learning is to generate negative samples from a large sample set to contrast with positive samples, for learning better encoding of the data. These negative samples often follow a softmax distribution which are dynamically updated during the training process. However, sampling from this distribution is non-trivial due to the high computational costs in computing the partition function.  
In this paper, we propose an \underline{E}fficient \underline{M}arkov \underline{C}hain Monte Carlo negative sampling method for \underline{C}ontrastive learning ({\algname}). We follow the global contrastive learning loss as introduced in \cite{yuan2022provable}, and propose {\algname} which utilizes an adaptive Metropolis-Hastings subroutine to generate hardness-aware negative samples in an online fashion during the optimization. 
We prove that {\algname} finds an $\mathcal{O}(1/\sqrt{T})$-stationary point of the global contrastive loss in $T$ iterations.
Compared to prior works, {\algname} is the first algorithm that exhibits global convergence (to stationarity) regardless of the choice of batch size while exhibiting low computation and memory cost. 
Numerical experiments validate that {\algname} is effective with small batch training and achieves comparable or better performance than baseline algorithms. We report the results for pre-training image encoders on {\tt STL-10} and {\tt Imagenet-100}.
\end{abstract}

\section{Introduction}
Contrastive representation learning has been instrumental in self-supervised learning for large-scale pre-training of foundation models \cite{radford2021learning,cherti2023reproducible} as well as in the fine-tuning stage on downstream tasks \cite{xiong2020approximate,lindgren2021efficient}.
It helps encode real-world data into low-dimensional feature vectors that abstract the important attributes about the data, and generalize well outside of the training distribution. More recently, contrastive learning with multi-modal data has helped embed different data modalities into the same feature space \cite{li2023multimodal}, such as the studies with visual-language models \cite{radford2021learning,alayrac2022flamingo,cherti2023reproducible} and document understanding \cite{xu2020layoutlmv2,lee2023formnetv2}.

Contrastive learning uses pairwise comparison of representations in the training objective, with the goal of learning representations of data where positive pairs are drawn closer while negative pairs move apart in the representation space. It is well known that generating a large dataset of pairwise samples such as image-text pairs of the same semantics costs much lower than manual labeling, e.g., the WebImageText dataset used for training CLIP originates from Wikipedia articles \cite{radford2021learning}. While there have been many studies about training objectives for contrastive learning, optimizing the contrastive loss efficiently remains an open-problem as current optimization methods critically rely on a large batch size to maintain the quality of the negative sample distribution.


Formally, we aim to train models $\phi: \mathcal{X} \rightarrow  \mathbb{R}^d$ and $\psi: \mathcal{Y} \rightarrow \mathbb{R}^d$ that are parameterized by $\theta \in \mathbb{R}^p$. These models encode data from the input space ${\cal X}, {\cal Y}$ of potentially two modalities into a vector space. Given $\data_{X} \subseteq {\cal X}$, $\data_{Y} \subseteq {\cal Y}$, we define a distribution of positive pairs $\dpos$ such that ${\rm supp}(\dpos) \subseteq \data_{X} \times \data_{Y}$ and a negative set $\dneg(x) \subseteq \data_Y$ for every data point $x \in {\bf D}_{X}$. We denote the dataset constructed in this manner by the tuple $(\dpos, \dneg(\cdot))$. 

This paper aims to minimize the following global contrastive loss \cite{lindgren2021efficient, yuan2022provable} with inverse temperature $\beta > 0$:
\begin{align}
    \min_{\theta \in \mathbb{R}^p} & \, \mathcal{L}(\theta) \label{eq:gcl}\\
    &\hspace{-.3cm} := \mathop{\mathbb{E}}_{(x,y) \sim \dpos}\left[-\log \frac{\exp(\beta \phi(x;\theta)^\top \psi(y;\theta))}{ {\displaystyle \smath{\sum_{z \in \dneg(x)}}} \exp(\beta \phi(x;\theta)^\top \psi(z;\theta))}\right] \notag
\end{align}
also see \cite{mikolov2013distributed} that proposed a similar loss to \eqref{eq:gcl}. Assume $m := |{\rm supp}(\dx)| < \infty$ where $\dx$ is the marginal distribution of $x$ on $\dpos$. Moreover, we let $\mneg := |\dneg(x)| < \infty$ for all $x \in \data_X$\footnote{The results of this paper can be extended to the case where the negative sample size $|\dneg(x)|$ is uneven.}.

The global contrastive loss \eqref{eq:gcl} aims at finding the feature encoders $\phi^\star, \psi^\star$ that map data into $\mathbb{R}^d$ with maximized similarity $\phi^\star(x)^\top \psi^\star(y)$ between positive data pair $(x,y) \sim \dpos$ and minimized similarity $\phi^\star(x)^\top \psi^\star(z)$ between negative data pair $(x,z)$, $z \in \dneg(x)$. In other words, the loss maximizes the probability of observing a positive sample $y$ under the context $x$ among the other negative samples.


We notice that a highly related loss function design to \eqref{eq:gcl} is the InfoNCE loss  \cite{logeswaran2018efficient,oord2018representation,chen2020simple}, which is defined by: 
\vspace{-.5cm}

{\small \begin{align} 
&{\cal L}_{\rm NCE}(\theta; B)\label{eq:simclr_loss} \\
    &=\mathop{\mathbb{E}}_{(x,y)\sim\dpos} \mathop{\mathbb{E}}_{{\bf Z} \sim {\cal D}_{\rm neg}(x; B)} \left[ - \log \frac{\exp(\beta \phi(x;\theta)^\top \psi(y;\theta) )}{ {\displaystyle \smath{\sum_{z \in {\bf Z}}} \exp(\beta \phi(x;\theta)^\top \psi(z;\theta) ) }}\right] \notag
\end{align}}such that ${\cal D}_{\rm neg}(x; B)$ is a distribution of all subsets ${\bf Z} \subseteq \dneg(x)$ with $|{\bf Z}| = B$. 
In fact, the global contrastive loss \eqref{eq:gcl} is an upper bound of the InfoNCE loss \eqref{eq:simclr_loss} since 
\begin{align}
    &\mathop{\mathbb{E}}_{{\bf Z} \sim {\cal D}_{\rm neg}(x; B)} \left[\log \sum_{z \in {\bf Z}} \exp(\beta~\phi(x;\theta)^\top \psi(z;\theta)) \right] \\
    &\leq \log \sum_{ z \in \dneg(x)} \exp(\beta~\phi(x;\theta)^\top \psi(z;\theta)), \notag
\end{align}
for any $x \in {\rm supp}(\dpos), \theta \in \mathbb{R}^p$. As the equality holds when $B = \mneg$, we have ${\cal L}_{\rm NCE}(\theta; \mneg) = {\cal L}(\theta)$.

It was demonstrated that minimizing \eqref{eq:simclr_loss} with $B \gg 1$ produces encodings with interpretable geometric and semantic properties \cite{wang2020understanding,robinson2020contrastive,zimmermann2021contrastive}. At the same time, the convergence of pre-training foundation models on large scale dataset happens only when using a large enough batch size $B$ \cite{radford2021learning}. Such evidence points towards the success of models that are trained to minimize ${\cal L}_{\rm NCE} (\theta; B )$ of large negative batch size $B$.
We believe that minimizing ${\cal L}(\theta)$, which yields the limiting upper bound to \eqref{eq:simclr_loss}, can lead to a better performance for contrastive learning. 

\begin{table*}[htpb] 
\begin{minipage}{\textwidth}
\centering
\resizebox{\linewidth}{!}{
\begin{tabular}{l|ccc} 
\toprule
\bfseries{Algorithms}& 
\bfseries{Convergence Error}& 
\bfseries{Memory}&
\bfseries{Computation} \\ \midrule 
\begin{tabular}{@{}l@{}} {SimCLR} \\ {\cite{chen2020simple}}\end{tabular} & ${\cal O}\left(\frac{\beta^2 }{\sqrt{T}} + \frac{\mneg^2 \exp(4\beta) \sigma^2}{B} \right)$${}^\star$ & ${\cal O}(BM_\phi + Bd)$ & ${\cal O}(BC_\phi + B^2d)$  \\[1em]
\begin{tabular}{@{}l@{}} {Negative Cache} \\ {\cite{lindgren2021efficient}}\end{tabular} & ${\cal O}\left(\frac{\beta}{\sqrt{T}} + \frac{\beta^6}{\rho^2 T} \right)$${}^\mathsection$ & ${\cal O}(M_\phi + \mneg d + d)$ & ${\cal O}(C_\phi + \mneg d + \rho m C_\phi)$ \\[1em]
\begin{tabular}{@{}l@{}} {SogCLR} \\ {\cite{yuan2022provable}}\end{tabular} & ${\cal O}\left( \frac{\exp(6\beta)}{\sqrt{BT}} + \frac{\mneg^2 \exp(4\beta)\sqrt{m}\sigma^2}{B\sqrt{T}} + {\mneg^2 \exp(4\beta)\upsilon^2} \right)$${}^\dagger$ & ${\cal O}(BM_\phi + m + Bd)$ & ${\cal O}(BC_\phi + B^2d)$  \\ [1em]
\cellcolor{gray!10}\bfseries\algname~(Ours, $R=2$) & \cellcolor{gray!10}${\cal O}\left({\frac{\beta}{\sqrt{T}} + \frac{m^2 \mneg^2 \exp(6c^2\beta)\beta^3 \sigma^2  }{B \sqrt{T}}} \right)$ & \cellcolor{gray!10}${\cal O}(BM_\phi + m + Bd)$ & \cellcolor{gray!10}${\cal O}(BC_\phi + B^2d)$ \\  
\cellcolor{gray!10}\bfseries\algname~(Ours) & \cellcolor{gray!10}${\cal O}\left({\frac{\beta}{\sqrt{T}} + \frac{2^R m^2 \mneg^2 \exp(6c^2\beta)\beta^3 \sigma^2  }{B R^2 \sqrt{T}}} \right)$ & \cellcolor{gray!10}${\cal O}(BM_\phi + m + Bd)$ & \cellcolor{gray!10}${\cal O}(BC_\phi + B^2d + BR)$ \\ \bottomrule 
\end{tabular}
}
\caption{\label{table:compl} The second column shows the upper bound on convergence error $T^{-1} \sum_{t=0}^{T-1}\mathbb{E}\left[\| \nabla_\theta {\cal L}(\theta_t) \|_2^2\right]$. The last two columns show the memory/computation requirement per iteration of the algorithms. $C_\phi$ (resp. $M_\phi$) denotes the computational (resp. memory) cost to compute the feature vector $\phi(x;\theta)$ or $\psi(x;\theta)$. ${}^\star$From (Theorem 1, \citealt{yuan2022provable}). ${}^\mathsection$The analysis shown in \cite{lindgren2021efficient} only consider the case when batch size $B=1$. ${}^\dagger$SogCLR only converge to an ${\cal O}(\upsilon^2)$-stationary solution, where $\upsilon$ is the worst-case feature heterogeneity error satisfying $\sup_{\theta \in \mathbb{R}^p}\mathop{\mathbb{E}}_{(x,y) \sim \dpos} \mathop{\mathbb{E}}_{z \sim {\rm Uniform}( \dneg(x))} \left[ |\phi(x;\theta)^\top \psi(z;\theta) - \phi(y;\theta)^\top \psi(z;\theta)| \right] \leq \upsilon^2$.
}
\end{minipage}
\end{table*}

\textbf{Challenges in Optimizing \eqref{eq:gcl}.} We notice that the contrastive loss gradient is given by:
\begin{align}
    &\nabla {\cal L}(\theta) = \mathop{\mathbb{E}}_{(x,y) \sim \dpos }\Big[-\beta ~\nabla_\theta(\phi(x;\theta)^\top \psi(y;\theta)) \Big]  \label{eq:gcl_grad_sum} \\
    & +\mathop{\mathbb{E}}_{(x,y) \sim \dpos } \Big[ \beta \sum_{z \in \dneg(x)} p_{x,\theta}(z)  \nabla_\theta(\phi(x;\theta)^\top \psi(z;\theta)) \Big] \notag \\
    & \equiv \nabla {\cal L}_{\rm pos}( \theta ) + \nabla {\cal L}_{\rm neg}( \theta ) , \notag
\end{align}
with the softmax distribution: 
\begin{equation}
    p_{x,\theta}(z) = \frac{\exp(\beta ~ \phi(x;\theta)^\top \psi(z;\theta)) }{\sum_{z' \in \dneg(x) } \exp(\beta ~ \phi(x;\theta)^\top \psi(z';\theta))}. \label{eq:def_pij}
\end{equation}
The challenge of optimizing ${\cal L}(\theta)$ lies in the overwhelming complexity to compute or approximate $\nabla {\cal L}_{\rm neg}(\theta)$ since $\dneg(x)$ often spans a large dataset, e.g., $\mneg = 8.8 \times 10^6$ for the {\tt MS MARCO} dataset \cite{bajaj2016ms}. 

To this end, SimCLR \cite{chen2020simple} proposed to replace $\dneg(x)$ in \eqref{eq:gcl_grad_sum} by a randomly selected \emph{negative batch} ${\bf Z}$ of $B$ randomly augmented images, i.e., using the gradient of InfoNCE loss \eqref{eq:simclr_loss}. While achieving reasonable performance in certain scenarios, SimCLR requires a large batch size $B$ on large-scale dataset training. For example, it requires up to $B=32768$ negative samples per iteration in training CLIP \cite{radford2021learning}, making it impossible to train such models in non-commercial data center \cite{cherti2023reproducible}.
We note that there exists a number of contrastive learning tricks to avoid the problem of estimating $\nabla {\cal L}_{\rm neg}( \theta )$. For example, \cite{robinson2020contrastive} imposes a hardness-aware distribution on ${\cal D}_{\rm neg}(x;B)$ of the InfoNCE loss. \cite{he2020momentum} utilizes a momentum mechanism to prevent feature collapse. \cite{grill2020bootstrap,zbontar2021barlow,bardes2021vicreg} use alternative loss functions for contrastive learning, such as redundancy reduction loss and covariance regularization. See \cite{balestriero2023cookbook} for a comprehensive overview on different contrastive learning methods.

An alternative approach is to consider the estimation of $\nabla {\cal L}_{\rm neg}(\theta)$ through \emph{negative sampling}. This approach is motivated from the observation that \eqref{eq:def_pij} is a probability mass function of a softmax distribution and thus the summation $\sum_{ z \in \dneg(x) } p_{x,\theta}(z) \nabla_\theta(\phi(x;\theta)^\top \psi(z;\theta))$ in $\nabla {\cal L}_{\rm neg}(\theta)$ is equivalent to an expectation taken w.r.t.~the softmax distribution. Subsequently, it can be approximated through \emph{sampling}. In particular, if one draws $(x,y) \sim \dpos$, $z \sim \px$ where  
\begin{equation} \label{eq:px_def}
\px \equiv ( p_{x,\theta} (z) )_{z \in \dneg(x)} ,
\end{equation} 
then $\beta \nabla_\theta(\phi(x;\theta)^\top \psi(z;\theta))$ is an unbiased estimate of $\nabla {\cal L}_{\rm neg}(\theta)$. 

However, sampling from $\px$ remains a highly non-trivial task due to the complexity in computing the partition function (i.e., denominator) in the softmax distribution \eqref{eq:def_pij}. Prior works have proposed remedies to the sampling problem. The \emph{negative cache} algorithm \cite{lindgren2021efficient} stores features $\{\psi(z; \theta_{\tau})\}_{z \in \dneg(x)}$ from previous iterations and apply the Gumbel-max trick for sampling from an approximate $\px$. Though it is proven to converge to a stationary point of \eqref{eq:gcl}, the algorithm suffers from high computation and memory complexity. 
Under a similar purpose, SogCLR \cite{yuan2022provable} proposed a running mean estimator for the normalization sum in $\px$. While SogCLR corrects the gradient bias without a significant sampling nor computation overhead, the algorithm is only guaranteed to converge to a neighborhood of stationary points of \eqref{eq:gcl} with a non-vanishing error. A comparison of the above works is in Table~\ref{table:compl}.

\begin{figure}[t]
    \centering \vspace{-0.15cm}
    \includegraphics[width=0.375\textwidth]{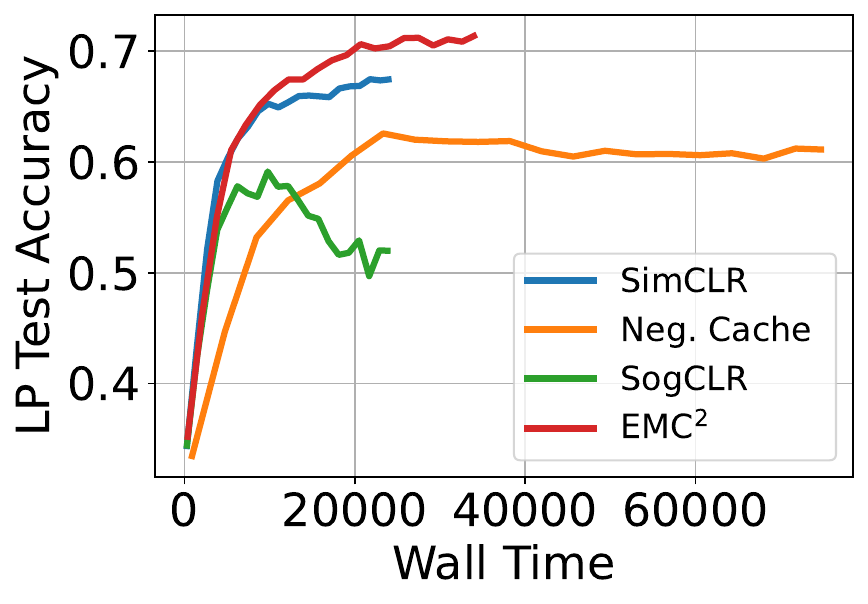} \vspace{-.2cm}
    \caption{Training 100 epochs on {\tt STL-10} with ResNet-18 using batch size $b = 32$. Horizontal axis is relative to the wall-clock training time in seconds. 
    }\vspace{-.2cm}
    \label{fig:stl10_teaser}
\end{figure}
\textbf{Our Contributions.} We propose \algname, an \underline{E}fficient \underline{M}arkov \underline{C}hain Monte Carlo negative sampling method for \underline{C}ontrastive Learning. Our method departs from the existing approaches since {\algname} directly tracks $\px$ to generate negative samples for $\nabla {\cal L}_{\rm neg}(\theta)$. More specifically,
\begin{itemize}[leftmargin=*, topsep=0mm, itemsep=0mm]
    \item The {\algname} utilizes a Metropolis-Hasting (M-H) algorithm specialized for negative sampling from \eqref{eq:def_pij}. Moreover, the samples from our M-H algorithm \emph{dynamically adjust} to the evolution of the stochastic gradient (SGD) iterates. This results in a state-dependent SGD scheme that enjoys low memory and computation complexity per iteration.
    \item We characterize the finite-time convergence rate of {\algname}. We show that it converges in expectation to a stationary point satisfying $\mathbb{E} [ \| \nabla {\cal L} ( \theta_t ) \|^2 ] = {\cal O}( 1 / \sqrt{T} )$ for some $t \in \{1,...,T\}$, where $T$ is the number of iterations. Moreover, the latter is neither affected by small batch size, nor the burn-in period with the M-H algorithm; see Table~\ref{table:compl}. 
    \item Our convergence analysis involves a non-trivial adaptation of the generic result for biased stochastic approximation scheme in \cite{karimi2019non}. Specifically, we prove that the $\theta$ dependent kernel which induces the MCMC's Markov chain is ergodic, and is Lipschitz w.r.t.~$\theta$.
\end{itemize}
Figure~\ref{fig:stl10_teaser} previews the performance of {\algname} for pre-training the image encoder on the {\tt STL-10} dataset. Observe that when training with a small batch size, {\algname} is around 2x faster than SimCLR and SogCLR, and 3x faster than Negative Caching.
The rest of this paper is organized as follows. Section \ref{sec:method} develops the {\algname} algorithm by showing how to combine M-H sampling in an online manner with SGD updates. Section \ref{sec:conv} presents the main convergence analysis results. Finally, Section \ref{sec:ex} shows the numerical experiments to corroborate our claims on the efficacy of {\algname}.

\textbf{Related Works.} We remark that existing works have considered using Markovian samples that is not i.i.d.~in SGD optimization. General convergence analysis results have appeared in \cite{sun2018markov, doan2022finite} for convex and non-convex optimization with a homogeneous Markov chain, and \cite{atchade2017perturbed, karimi2019non} with controlled Markov chain. Example applications include policy evaluation in reinforcement learning \cite{baxter2001infinite, bhandari2018finite, srikant2019finite}, Bayesian optimization for maximum likelihood \cite{de2021efficient}, expectation maximization with stochastic samples \cite{kuhn2004coupling}. 

For tasks related to contrastive learning, prior works have proposed using tree indexed structure to accelerate sampling \cite{monath2023improving}, and using the graph structure for graph representation learning \cite{yang2020understanding}. However, they only proved that the bias of the gradient estimator is bounded and lack convergence analysis for the overall learning algorithm. We remark that there are earlier works on estimating the cross-entropy loss over a large set of classes \cite{vembu2012probabilistic}, e.g., by sub-sampling the classes based on kernel methods \cite{blanc2018adaptive} and random Fourier features \cite{rawat2019sampled}. In comparison, our work introduces the technique to contrastive learning, and provide a comprehensive theoretical and empirical analysis on its efficacy and convergence properties.

\section{Our Proposed Method: \algname} \label{sec:method}
This section develops {\algname} for optimizing the global contrastive loss ${\cal L}( \theta )$ in \eqref{eq:gcl}. To simplify notation, we define\begin{equation}
    H( x, y ; \theta ) := \nabla_\theta \big[ \phi(x;\theta)^\top \psi(y;\theta) \big] 
\end{equation}
as the gradient of the sample pair $(x,y) \in {\cal X} \times {\cal Y}$.
Recall from \eqref{eq:gcl_grad_sum} that the population gradient $\nabla {\cal L}( \theta )$ is composed of two terms: (i) $\nabla {\cal L}_{\rm pos}( \theta )$ maximizes the correlation between positive sample pairs, (ii) $\nabla {\cal L}_{\rm neg}( \theta )$ minimizes the correlation between \emph{negative} sample pairs. 

To apply SGD on \eqref{eq:gcl}, with a positive sample pair drawn uniformly as $(x,y) \sim \dpos$, the vector $- \beta H( x,y ; \theta )$ yields an \emph{unbiased} estimate for $\nabla {\cal L}_{\rm pos}( \theta )$. 
Our challenge lies in obtaining an \emph{unbiased} estimate for the negative sample gradient $\nabla {\cal L}_{\rm neg}( \theta )$.
Observe that 
\begin{equation}
    \nabla {\cal L}_{\rm neg}( \theta ) = \hspace{-.0cm} \mathop{\mathbb{E}}_{\begin{subarray}{c} 
    (x,y) \sim \dpos,~
    z \sim \px \end{subarray} } 
    \Big[ \beta H( x,z ; \theta ) \Big] \label{eq:gcl_grad_pij}
\end{equation}
for $\px$ defined in \eqref{eq:px_def}. Compared to the case of $\nabla {\cal L}_{\rm pos}( \theta )$, the distribution for the tuple $(x,z)$ in the expectation above depends on $\theta$. It follows a softmax distribution \eqref{eq:def_pij} with a large summation in the denominator that is difficult, if not impossible, to evaluate when the number of negative samples is large. Furthermore, the distribution also depends on $\theta$ and has to be updated \emph{dynamically} as we optimize $\theta$. 

We refer to the task of estimating \eqref{eq:gcl_grad_pij} as the \emph{negative sampling} problem.
To improve memory consumption and computation cost, our idea is to develop an MCMC negative sampling method that generates the desired samples in an online manner. The resultant SGD method for \eqref{eq:gcl} is then treated as a stochastic approximation scheme with state-dependent samples coming from a \emph{controlled Markov chain}.

\subsection{Negative Sampling via MCMC}
The MCMC scheme \citep{robert1999monte} is a classical yet powerful method for generating samples from an arbitrary distribution $\pi$. We focus on the Metropolis-Hastings (M-H) algorithm \cite{chib1995understanding} due to its simplicity. In a nutshell, the algorithm generates new candidate samples from a uniform distribution and adjusts the frequency of samples by a reject/accept mechanism to match the target distribution $\pi$. It induces a Markov chain whose stationary distribution is the target distribution. 

For the negative sampling problem in \eqref{eq:gcl_grad_pij}, with $(x,y) \sim \dpos$, we generate samples $z \sim \px$ by running the M-H algorithm. In particular, we assign a state $Z_i \in [ \mneg ]$ for each $i \in [m]$ that corresponds to a positive sample $x_i \in \data_{X}$. We generate a candidate sample $Z_i' \sim {\rm Uniform}( [ \mneg ] )$, to be accepted as the new sample $Z_i^+$ with probability: 
\begin{equation} \label{eq:mh_accept}
Q_{x_i,\theta}( Z_i' ,Z_i ) = \frac{ p_{x_i,\theta}(Z_i')}{p_{x_i,\theta}( Z_i )} = \frac{\exp(\beta ~ \phi(x_i;\theta)^\top \psi(Z_i';\theta))}{\exp(\beta ~ \phi(x_i;\theta)^\top \psi(Z_i;\theta))} ,
\end{equation}
where $Z_i$ is the old sample; if the sample $Z_i'$ is rejected, then $Z_i^+ = Z_i$. 
The above procedure induces a Markov chain $\cdots \to Z_i \to Z_i^+ \to \cdots$ whose stationary distribution coincides with ${\cal P}_{ x_i , \theta }$. Together with $(x_i, y_i) \equiv (x,y) \sim \dpos$, this allows us to construct an unbiased estimate for $\nabla {\cal L}_{\rm neg}(\theta)$ as $\beta H( x_i, {Z}_i^\infty; \theta )$, where ${Z}_i^\infty$ is obtained after running the M-H algorithm for a certain number of steps.

We note that evaluating \eqref{eq:mh_accept} does not require us to compute the partition function in the denominator of \eqref{eq:def_pij}. As such, the computation complexity of evaluating \eqref{eq:mh_accept} is only $2 C_\phi$. On the other hand, we require storing the state $Z_i$ for each positive sample $x_i \in \data_X$, which results in a memory cost of $m$ integers. The latter needs not be updated for every iteration.
Compared to the negative caching algorithm \cite{lindgren2021efficient} which requires a cache memory consumption of $\mneg d$ real numbers for storing the feature vectors $\psi( z; \theta )$ for all $z \in \dneg(x)$, and a computation cost of $\rho \mneg C_\phi$ for updating a $\rho$-fraction of cache, the MCMC scheme enjoys a lower memory cost and computation complexity.\vspace{.1cm} 

{\bf Single State MCMC vs Multi-State MCMC. }
Alternatively, one can apply the M-H algorithm to generate samples from the \emph{joint distribution} of $(x,z)$ in \eqref{eq:gcl_grad_pij} using a single state MCMC. However, the corresponding reject/accept ratio [cf.~\eqref{eq:mh_accept}] involves the normalization constant $G(x,\theta) = \sum_{z \in \dneg(x)} \phi(x;\theta)^\top \psi(z;\theta)$ from \eqref{eq:def_pij}, for which $G(x,\theta)$ and $G(x', \theta)$ have to be re-evaluated when transitioning from $(x, \cdot)$ to $(x', \cdot)$. Implementing such a scheme will require a $2 m_{\rm neg} C_\phi$ computation complexity, or require approximating the normalization constant as a constant that is independent of $x$, both of which are encouraging the use of a multi-state Markov chain as illustrated in \eqref{eq:mh_accept}.

\subsection{State-dependent SGD Algorithm}
As the target distribution in \eqref{eq:gcl_grad_pij} depends on $\theta$, applying SGD with the classical MCMC scheme or M-H algorithm requires first \emph{freezing} $\theta$ and then simulating the Markov chain for a considerable amount of time \emph{prior to} forming the stochastic gradient. The latter is known as the \emph{burn-in} period for MCMC and incurs additional complexity for every SGD iteration.

We propose to adjust the Markov chains in an online manner as $\theta$ is updated, i.e., the method shall not maintain a long burn-in period. 
In particular, at each SGD iteration, the M-H updates are executed with initialization given by the state stored previously. Meanwhile, the Markov chains with the reject/accept ratio \eqref{eq:mh_accept} are controlled by the current $\theta$. 

\begin{algorithm}[t] 
\caption{Efficient MCMC Negative Sampling Method for Contrastive Learning (\algname)} 
\begin{algorithmic}[1]\label{alg:multi_mcmc}
\STATE {\bfseries input:} Iteration number $T$, batch size $B$, negative batch size $R$, burn-in period $P < R$, Markov chain state initialization $\{Z_j\}_{j=1}^{m}$ and step size $\gamma$.
\FOR{$t = 0, ..., T-1$}
    \STATE Draw a mini-batch $[ (x_{i_1^{(t)}}, y_{i_1^{(t)}}), ..., (x_{i_B^{(t)}}, y_{i_B^{(t)}}) ] \sim \dpos$, with indices $\{ i_1^{(t)}, ..., i_B^{(t)} \} \subseteq [m]$.
    \FOR{$k = 1, ..., B$; $r = 0, ..., R-1$}
        \STATE Draw negative sample $Z'_{i_k^{(t)}} \sim {\rm Unif}(\dneg(x_{i_k^{(t)}}))$.
        \STATE Update the Markov chain state $Z_{i_k^{(t)}}$ by \vspace{-.2cm}
        \[ \hspace{-.8cm} Z_{i_k^{(t)}} \leftarrow
            \begin{cases}
                Z'_{i_k^{(t)}} & \text{w.p.} ~ \min\left\{ 1, ~ Q_{x_{i_k^{(t)}},\theta_t}(Z'_{i_k^{(t)}},Z_{i_k^{(t)}}) \right\} , \\
                Z_{i_k^{(t)}} &\text{otherwise},
            \end{cases} \] 
            where $Q_{x, \theta}(Z',Z)$ is defined in \eqref{eq:mh_accept}.
        \STATE If $r \geq P$, store the sample $\tilde{Z}_{i_k^{(t)}}^{(r)} = Z_{i_k^{(t)}}$.
    \ENDFOR
        \STATE Update the model $\theta$ using \eqref{eq:sgd}.
\ENDFOR
\end{algorithmic}
\end{algorithm}

Let $t$ denotes the SGD iteration index and $\theta_{t}$ is the corresponding model. Suppose that a mini-batch of $B$ positive samples are drawn $\{ x_{i_k^{(t)}}, y_{i_k^{(t)}} \}_{k=1}^B$ and $\{ \{ \tilde{Z}_{ i_k^{(t)} }^{(r)} \}_{r=0}^{R-1} \}_{k=1}^B$ are the sequences of samples generated by $R$ steps of the M-H algorithm, where the latter is initialized by the states $Z_{i_k^{(t)}}$ from the previous SGD iteration. Our idea is to update the model $\theta$ by the following recursion:
\begin{equation} \label{eq:sgd}
    \theta_{t+1} = \theta_t - \gamma {\cal H} \left( \bm{\xi}_{t+1} ; \theta_t \right) ,
\end{equation}
where $\bm{\xi}_{t+1} := \{ x_{i_k^{(t)}}, y_{i_k^{(t)}}, \{ \tilde{Z}_{ i_k^{(t)}}^{(r)} \}_{r=P}^{R-1} \}_{k=1}^B$ collects the samples used, $\gamma > 0$ is the SGD step size, and
\begin{align}
& {\cal H} \left( \bm{\xi}_{t+1} ; \theta_t \right) := - \frac{\beta}{B} \sum_{k=1}^B H( x_{i_k^{(t)}}, y_{i_k^{(t)}} ; \theta_t ) \label{eq:sg} \\
& \qquad + \frac{\beta}{B(R-P)} \sum_{k=1}^B \sum_{r=P}^{R-1} H( x_{i_k^{(t)}}, \tilde{Z}_{i_k^{(t)}}^{(r)} ; \theta_t ), \notag
\end{align}
where $P$ is an adjustable burn-in parameter.
The overall algorithm is summarized in Algorithm~\ref{alg:multi_mcmc}.

We observe that \eqref{eq:sgd} is different from a standard SGD algorithm as ${\cal H} ( \bm{\xi}_{t+1} ; \theta_t )$ corresponds to a \emph{biased} estimate for the gradient $\nabla {\cal L} ( \theta_t )$ of the contrastive learning loss \eqref{eq:gcl} in general. Instead, the algorithm belongs to the more general class of \emph{biased stochastic approximation} scheme \cite{karimi2019non} whose stochastic updates are driven by the Markov chain $\cdots \to \bm{\xi}_t \to \bm{\xi}_{t+1} \to \cdots$. Moreover, the Markov transition kernel is controlled by the current model $\theta_t$ that is updated simultaneously.

To reduce bias, a traditional approach is to consider $P \to \infty$, $R = P+1$ which will ensure that the M-H algorithm can generate its samples from the stationary distribution. However, as we shall demonstrate in Section ~\ref{sec:conv}, under suitable and verifiable conditions, the $\theta$-dependent Markov transition kernel is Lipschitz continuous and uniform geometrically ergodic. Subsequently, it ensures the convergence of \eqref{eq:sgd} towards a stationary point of \eqref{eq:gcl} for any $P,R$.

\section{Convergence Analysis} \label{sec:conv}
This section shows that {\algname} converges to a stationary point of \eqref{eq:gcl}.
Our analysis is organized as follows: first we characterize the mixing rate of the Markov chain $\cdots \to \bm{\xi}_t \to \bm{\xi}_{t+1} \to \cdots$ in \eqref{eq:sgd} and its smoothness property, then we analyze the convergence of the algorithm through studying the latter as a special case of the biased stochastic approximation scheme in \cite{karimi2019non}. Finally, we present our main results in Theorem~\ref{thm:main} and Corollary~\ref{cor:main}. 

\subsection{Analysis of MCMC Negative Sampling}
To enable the convergence of \eqref{eq:sgd} utilizing the Markov chain $\{ \bm{\xi}_{t} \}_{t \geq 0}$, intuitively it requires the Markov chain to (i) have a fast mixing time, and (ii) satisfy certain smoothness property as $\theta_t$ is gradually updated. To fix notation, we denote the state-dependent Markov transition kernel ${\tt P}_{ \theta }: \Xi \times \Xi \to \mathbb{R}_+$ such that $\bm{\xi}_{t+1} \sim {\tt P}_{ \theta_t } ( \xi_t , \cdot )$. For simplicity, we analyze the latter transition probability of $\bm{\xi}_{t+1}$ when $P = R-1$, i.e., the state dependent SGD update only takes the last sample in the M-H algorithm. Moreover, with a slight abuse of notation, we consider the collection of all hidden states $\bm{\xi}_{t+1} = 
\{ \tilde{Z}_i^{(R-1)} \}_{i=1}^m \in \Xi$
which does not include $x,y$. As the latter is drawn i.i.d., it does not affect the convergence analysis for \eqref{eq:sgd} in Section ~\ref{sec:ana_sgd}.

For our first task in showing the fast mixing of $\{ \xi_t \}_{t \geq 0}$, it suffices to show that the Markov chain induced by ${\tt P}_{ \theta }$ is \emph{geometrically ergodic} for any fixed $\theta$. To this end, we note that as the target distribution $\px$ belongs to the exponential family, applying (Theorem 2.1, \citealt{roberts1996geometric}) implies ergodicity. To formally state the result, we impose the following assumptions: 
\begin{assumption} \label{assm:bounded_embd}
    There exists $c > 0$ such that 
    \begin{equation}
        \max \{ \| \phi(x;\theta) \|,  \| \psi(x;\theta) \| \}  \leq c,~\forall~\theta \in \mathbb{R}^p,~x \in {\cal X}.
    \end{equation}
\end{assumption}
Assumption \ref{assm:bounded_embd} is common with $c \leq 1$ for contrastive learning problems \cite{chen2020simple,radford2021learning}. This effectively controls the behavior of $Q_{x,\theta}(z, Z)$ in the M-H algorithm \eqref{eq:mh_accept} and thus the transition kernel ${\tt P}_{\theta}$. The following lemma shows the geometric ergodicity property for the Markov chain:
\begin{lemma} \label{lemma:mcmc_rate}
    Under Assumption \ref{assm:bounded_embd}. For any $\theta \in \mathbb{R}^p$ and any initialization $\tilde{\bm{\xi}}_0 \in \Xi$, the Markov chain $\tilde{\bm{\xi}}_0 \to \tilde{\bm{\xi}}_1 \to \cdots$ induced by the transition kernel ${\tt P}_{\theta}$ converges geometrically to the stationary distribution $\pi_{{\bm x}, \theta}( {\bm z} ) = \prod_{i=1}^m 
    p_{x_i, \theta}(z_i)$, where ${\bm z} = (z_i)_{i=1}^m$. In particular, it holds 
    \begin{equation} \label{eq:rate_mh}
        \begin{split}
            \big| \mathbb{P}\big( \tilde{\bm{\xi}}_\tau = {\bm z} \big) - \pi_{{\bm x}, \theta}( {\bm z} ) \big| \leq \left( 1 - \frac{BR}{2 m\mneg \exp(2c^2 \beta)}\right)^\tau, \\
        \end{split}
    \end{equation}
    for any ${\bm z}$ and any $\tau \geq 0$.
\end{lemma}
See Appendix \ref{app:proof_mcmc_rate} for the proof, which relies on (Theorem 1.3, \citealt{mengersen1996rates}). Moreover, from \eqref{eq:rate_mh}, we observe that through mini-batch sampling among $\dpos$ and $\px$, the mixing rate improves by $BR$ when consuming $BR$ samples in {\algname}. 

Our second task consists of showing that the Markov transition kernel is smooth w.r.t.~$\theta$. We require the following condition the models $\phi, \psi$: 
\begin{assumption} \label{assm:smooth_kernel}
    There exists $L_P \ge 0$ such that for any $\theta, \theta' \in \mathbb{R}^p, (x, y) \in {\rm supp}(\dpos), z \in \dneg(x)$,
    \begin{equation}
        \| \phi(x; \theta)^\top \psi(z; \theta) - \phi(x; \theta')^\top \psi(z; \theta') \|\leq L_P \|\theta - \theta'\|.
    \end{equation}
\end{assumption}
Assumption \ref{assm:smooth_kernel} describes the Lipschitz condition of the similarity function $\phi(x;\theta)^\top \psi(z;\theta)$, which is important in establishing the Lipschitz condition of the transition kernel w.r.t. the model parameter $\theta$. For example, it can be satisfied as $L_P = Lc$ for $L$-Lipschitz functions $\phi, \psi$ w.r.t. $\theta$ with bounded norm as in Assumption \ref{assm:bounded_embd}.

The following lemma shows that the transition kernel ${\tt P}_{\theta}$ is Lipschitz continuous with respect to $\theta$, as follows:
\begin{samepage}
\begin{lemma} \label{lem:smooth}
Under Assumption \ref{assm:bounded_embd}, \ref{assm:smooth_kernel}. It holds:
\begin{equation} \label{eq:smooth_mh_bound}
    | {\tt P}_{\theta} ( \bm{\xi}, \bm{\xi}' ) - {\tt P}_{\theta'} ( \bm{\xi}, \bm{\xi}' ) | \leq 2^{R+1} B L_P \exp( 2c^2 \beta ) \beta \| \theta - \theta' \|,
\end{equation}
for any $\bm{\xi}, \bm{\xi}' \in \Xi$ and any $\theta, \theta' \in \mathbb{R}^p$.
\end{lemma}
\end{samepage}
The proof can be found in Appendix~\ref{app:smooth}. We remark that the dependence on $2^{R+1} B$ in the Lipschitz constant is conservative. We anticipate the effective Lipschitz constant for the transition kernel to be much smaller than in \eqref{eq:smooth_mh_bound}. For example, \eqref{eq:lose_bound} has applied a loose bound of $1$ to control the transition probability.

\subsection{Convergence to Stationary Point} \label{sec:ana_sgd}
Equipped with Lemma~\ref{lemma:mcmc_rate} \& \ref{lem:smooth}, we are ready to derive the convergence rate of Algorithm~\ref{alg:multi_mcmc} towards a stationary point of \eqref{eq:gcl}. Our idea is to treat \eqref{eq:sgd} as a biased stochastic approximation scheme and analyze the latter using the general convergence theories given in \cite{karimi2019non}. 

We require the following conditions:
\begin{assumption} \label{assm:smooth_sgrad}
    There exists $L_H \ge 0$ such that for any $\theta, \theta' \in \mathbb{R}^p, x \in \data_{X}, y \in \data_Y$,
    \begin{align}
      \| H(x,y; \theta) - H(x,y; \theta')  \|  &\leq L_H \|\theta - \theta'\| .
    \end{align}
\end{assumption}
\begin{assumption} \label{assm:bounded_var}
    There exists $\sigma \ge 0$ such that for any $\theta \in \mathbb{R}^p, (x, y) \in {\rm supp}(\dpos), z \in \dneg(x)$,
    \begin{align}
        \| H(x,y; \theta) - \mathbb{E}_{\dpos}[H(x,y; \theta)]  \|  &\leq \sigma, \\
        \| H(x,z; \theta) - \mathbb{E}_{\dpos, \px}[H(x,z; \theta)]  \|  &\leq \sigma.
    \end{align}
\end{assumption}
Assumptions \ref{assm:smooth_sgrad}, \ref{assm:bounded_var} impose a uniform bound on the smoothness condition of the similarity function and the variance of stochastic gradient. 
These assumptions are standard in the literature, e.g., they are also used in \cite{lindgren2021efficient, yuan2022provable}.

We present the main convergence result for {\algname}:
\begin{center}
\fbox{\begin{minipage}{.97\linewidth}{\vspace{.1cm}\begin{theorem} \label{thm:main}
    Under Assumptions \ref{assm:bounded_embd}, \ref{assm:smooth_kernel}, \ref{assm:smooth_sgrad}, \ref{assm:bounded_var}. For any $T \geq 1$, there exists a sufficiently small step size $\gamma > 0$ such that the iterate $\theta_t$ generated by {\algname} satisfies
    \begin{align}
        & \textstyle T^{-1} \sum_{t=0}^{T-1} \mathbb{E}[\| \nabla {\cal L} (\theta_t ) \|^2] \\
        &\leq \frac{2{\cal L}_{0,T}}{\gamma T} + \frac{ 24 \beta \sigma  m \mneg \exp(2c^2 \beta) }{\gamma B R T} + \gamma \beta^3 \sigma^2 \notag \\
        & \times \mathcal{O}\Big( \frac{ L_H m \mneg \exp(2c^2 \beta)}{BR}  + \frac{ L_P 2^{R} m^2 \mneg^2 \exp(6c^2 \beta)}{ B R^2 }  \Big) \notag 
    \end{align}
    where the above expectation is taken with respect to the randomness in the algorithm, and ${\cal L}_{0,T} = \mathbb{E}[{\cal L}(\theta_0) - {\cal L}(\theta_T)] \leq 4\beta$. 
\end{theorem}}\end{minipage}}
\end{center}

See Appendix \ref{app:proof_thm} for the proof. Theorem \ref{thm:main} implies that in expectation, {\algname} finds an ${\cal O}( \frac{1}{\gamma T} + \gamma )$ stationary solution to \eqref{eq:gcl}  in $T$ iterations, for any $T \geq 1$.  

For a sufficiently large $T$, setting $\gamma = 1/\sqrt{T}$ yields a global convergence rate of ${\cal O}(1/\sqrt{T})$ regardless of the batch size $B$ nor the burn-in period controlled by $P, R$, as observed in the following corollary:
\begin{corollary} \label{cor:main}
Under Assumptions \ref{assm:bounded_embd}, \ref{assm:smooth_kernel}, \ref{assm:smooth_sgrad}, \ref{assm:bounded_var}. For a sufficiently large $T$, choosing $\gamma = \frac{1}{\sqrt{T}}$, $R=2$, guarantees that the iterates generated by {\algname} satisfy:
    \begin{align}
        & \textstyle T^{-1} \sum_{t=0}^{T-1} \mathbb{E}[\| \nabla {\cal L}(\theta_t ) \|^2] \\
        &\leq \frac{8\beta}{\sqrt{T}} +\frac{ 12 \beta \sigma m \mneg \exp(2c^2 \beta) }{B^2 \sqrt{T}} +  \frac{\beta^3 \sigma^2 }{\sqrt{T}} \notag \\
        & \times \mathcal{O}\Big( \frac{ L_H m \mneg \exp(2c^2 \beta)}{B}  + \frac{ L_P m^2 \mneg^2 \exp(6c^2 \beta)}{ B }  \Big). \notag 
    \end{align}
\end{corollary}
We have presented a simplified version of the above result in Table~\ref{table:compl}.

\section{Numerical Experiments: Unimodal Pre-Training of Image Encoder} \label{sec:ex}
This section examines the performance of {\algname} for the contrastive learning task on training image encoders. 
We remark that although the analysis is only shown for the SGD optimizer with {\algname}, the stochastic gradient approximation in \eqref{eq:sg} can be applied to other popular first-order optimizers such as Adam.

Our experiments consider the task of pre-training self-supervised image encoder similar to \cite{chen2020simple}. This task is described by an instance of \eqref{eq:gcl} with unimodal encoders, i.e., $\phi(\cdot; \theta) \equiv \psi( \cdot; \theta )$ for any $\theta \in \mathbb{R}^p$. Moreover, the self-supervised dataset is specified with a set of augmented images. Let ${\cal A}$ be a set of image augmentation operators $g: {\cal X} \to {\cal X}$, we specify $( \dpos, \dneg(\cdot) )$ as:
\begin{equation} \label{eq:image_dual_encoder}
    \begin{cases}
    \dpos = {\rm Uniform}(\{ (g(x), h(x)): x \in {\bf D};~ g, h \in \mathcal{A}\}), \\
    \dneg(h(x)) = \{ g(y): y\neq x, y \in \data; g,h \in \mathcal{A} \}.
\end{cases}
\end{equation}
For the dataset $\data \subseteq {\cal X}$, the positive pairs are set of the random augmentations of the same image, while for $x \in \data$, its negative pairs are set to be the whole dataset with augmentation except that of $x$.

\begin{figure}[t]
    \centering
    \includegraphics[width=0.375\textwidth]{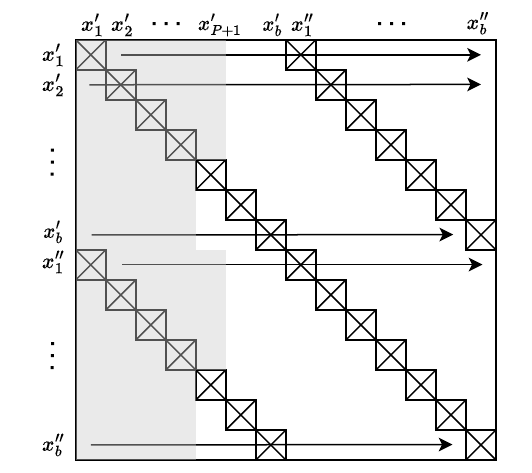}\vspace{-.2cm}
    \caption{Illustration of mini-batch MCMC sampling with 2 augmentations $(x_i', x_i'')$ of each image $x_i$. Each horizontal arrow represents a distribution tracked by one Markov chain and the direction of M-H reject/accept step. Shaded area represents the samples used for burn-in with burn-in period $P<2b-2$. Crossed-out diagonals are not regarded as negative samples.
    }
    \label{fig:batch_mcmc_illus}
\end{figure}

\subsection{Implementation of {\algname} }

\textbf{Image Augmentation.} We follow previous studies on applying input augmentations for contrastive learning on image data \cite{xie2020unsupervised} which would lead to samples from a distribution $\dpos$ with infinite support. Despite the fact that {\algname} is designed to track the Markov chain state for each image in $x_1, ..., x_m$, in practice we approximate Algorithm \ref{alg:multi_mcmc} by maintaining a Markov chain state for each set of $\{g(x_1): g\in {\cal A} \}, ..., \{g(x_m): g\in {\cal A} \}$\footnote{When the state $Z_i$ is referred outside the iteration it was sampled, it points to the image $x_i$ to avoid storing the augmented image $g(x_i)$ in memory.}. This approximate algorithm remains practical and performant as verified empirically in Section \ref{sec:finite_aug}.

\noindent \textbf{Mini-batch Sampling.} As proposed by \cite{chen2020simple}, it is efficient to draw a mini-batch and use the in-batch samples as negative samples. We can implement M-H reject/accept steps on top of the same mini-batch sampling scheme with $R=B-2$. In Figure \ref{fig:batch_mcmc_illus}, we illustrate a similarity matrix formed by the matrix product ${\bf V}^\top {\bf V}$ such that ${\bf V} = [\phi(x_1'), ...,\phi(x_b') , \phi(x_1''), ..., \phi(x_b'')] \in \mathbb{R}^{d \times 2b}$ contains the feature vectors of a mini-batch of size $B=2b$ with two random augmentations $(x_i', x_i'')$ of the same image $x_i$, $ i \in [b]$. {\algname} can regard the in-batch samples $\{ x_1', ..., x_b', x_1'', ..., x_b''\} \backslash \{x_j', x_j''\}$ as negative samples from the prior distribution ${\rm Uniform}(\dneg(x_j))$ and perform M-H acceptance/rejection steps over this negative set. As illustrated in Figure \ref{fig:batch_mcmc_illus}, the M-H algorithm can run in parallel across the multiple Markov chains. For the experiments, we consume burn-in samples from the same mini-batch without requiring extra samples.

\subsection{Dataset and Metrics}
We concentrate on two common datasets under this setup -- {\tt STL-10} and {\tt Imagenet-100}. 
We apply the Adam optimizer for training with {\tt STL-10} and the LARS optimizer with {\tt Imagenet-100}. Details of the datasets and hyperparameters used can be found in Appendix \ref{app:exp}.

To benchmark the performance of {\algname} and other algorithms, we report linear probe (LP) accuracy and 1-nearest-neighbor (1-NN) accuracy of the embeddings produced by the encoder, which are standard metrics for evaluating image encoders \cite{chen2020simple,kukleva2023temperature}. They correspond to the linear separability of features on a unit hypersphere \cite{wang2020understanding} and the local geometry of the learnt features.

Since the negative cache algorithm \cite{lindgren2021efficient} requires storing feature vectors of negatives, we are restricted to implementing on an alternative loss function with $\dneg(x) = \{y: y \in \data \backslash \{ x\} \}$, i.e., images without augmentation. Also, note that negative cache algorithm uses $B$ extra samples per iteration as the Gumbel-max negative sampling requires out-of-batch samples.

\begin{figure}[t]
    \centering
    \includegraphics[width=0.238\textwidth]{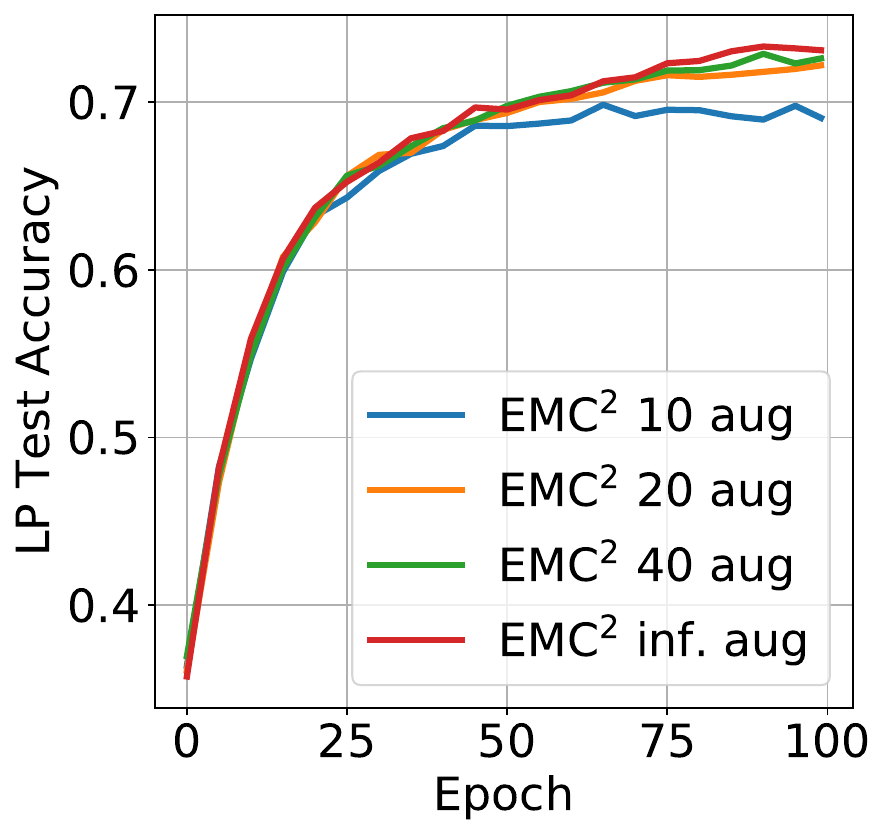}
    \includegraphics[width=0.238\textwidth]{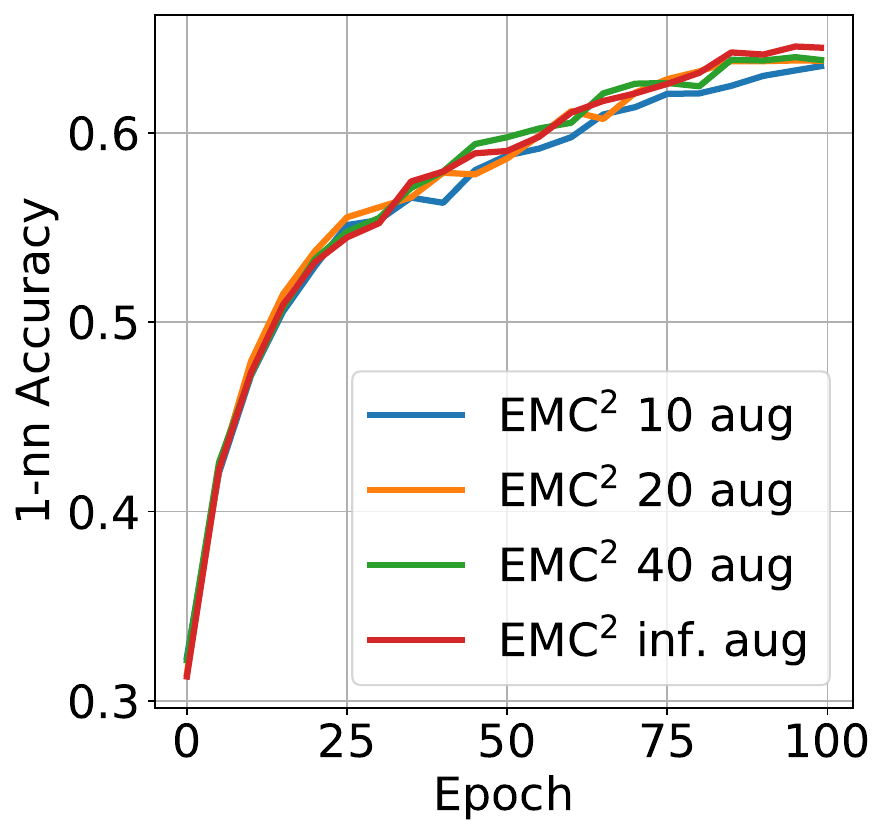}
    \caption{Comparison between different sizes of pre-augmented {\tt STL-10} with ResNet-18 and batch size $b = 256$. Horizontal axis is relative to the number of samples accessed.}
    \label{fig:stl10_finite_aug}
\end{figure}

\subsection{Effects of Image Augmentations} \label{sec:finite_aug} We first examine the effect of the number of pre-computed image augmentations, $|{\cal A}|$, on the performance of {\algname}. Note that as suggested in \cite{chen2020simple}, increasing $|{\cal A}|$ can lead to improved performance. We also compare a heuristic extension of {\algname} that effectively deploys an infinite number of augmentations: the augmented images are generated on-the-fly at every iteration and a single Markov chain state $Z_i$ is maintained for all augmentations of the same image $x_i$.


From Figure \ref{fig:stl10_finite_aug}, as the number of pre-computed augmentations increases, the performance approaches to that of the infinite augmentation algorithm. This illustrates that {\algname} adapts to random augmentations. In later experiments with image augmentations, we implement {\algname} with the infinite augmentation setting for the best performance.

\subsection{Effects of Burn-in Period}
\begin{figure}
    \centering \vspace{-0.15cm}
    \includegraphics[width=0.238\textwidth]{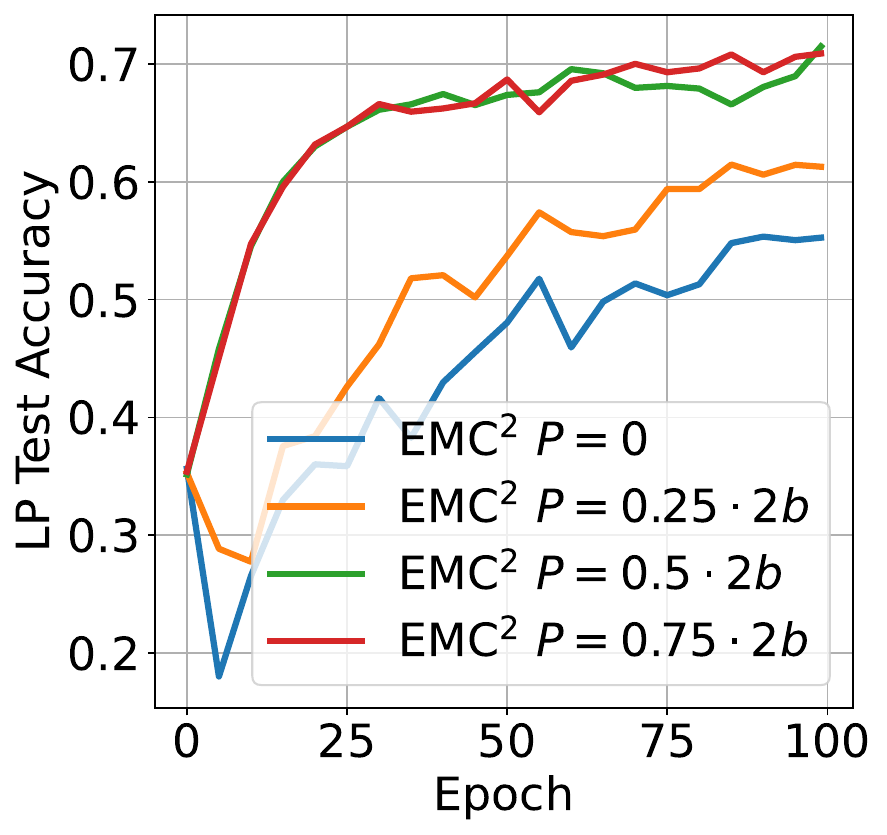}
    \includegraphics[width=0.238\textwidth]{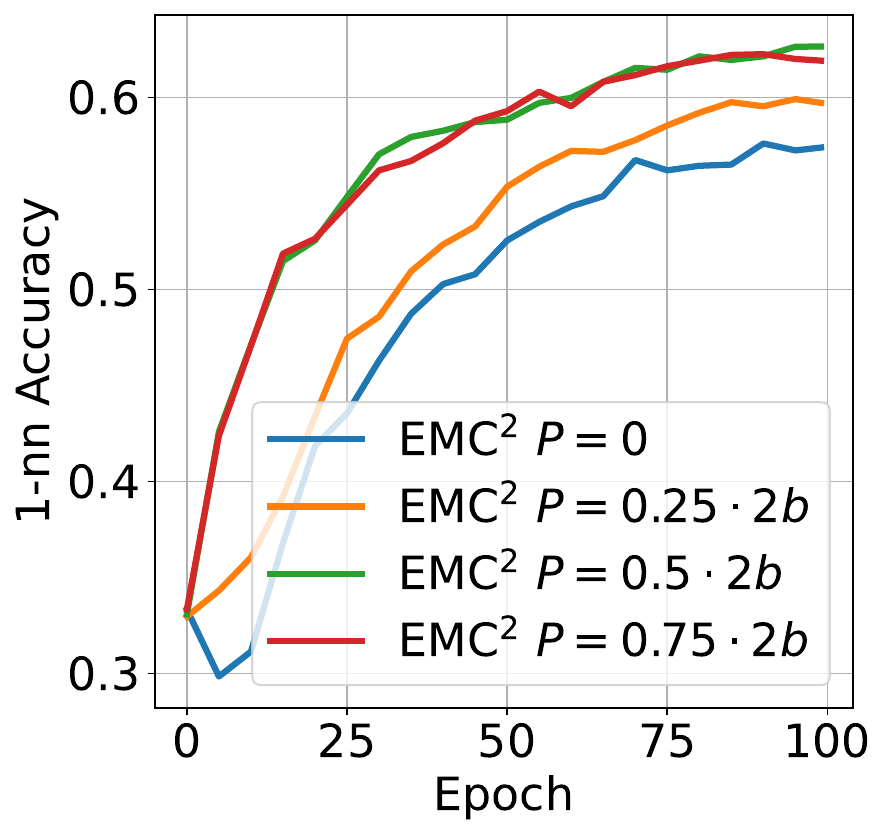}
    \caption{Comparison between different numbers of burn-in negative samples $P$ for each Markov chain state $Z_i$.}
    \label{fig:burn_in_ablation}
\end{figure}

For every positive pair $(x_i, y_i)$ in a mini-batch, {\algname} uses $P$ burn-in negative samples to warm-up the Markov chain state $Z_i$ of sample $x_i$. We study the effect of $P$ by comparing the performance for $P \in \{0, \frac{b}{2}, b, \frac{3}{2b} \}$. From Figure \ref{fig:burn_in_ablation}, we observe that the performance of {\algname} improves by increasing $P$, and the improvement stops at a certain threshold of around $P = b$. This amounts to the mechanism of Markov chain convergence where the negative samples after the burn-in period are more accurate to the true sample distribution in \eqref{eq:def_pij}.

\begin{figure}
    \centering \vspace{-0.15cm}
    \includegraphics[width=0.238\textwidth]{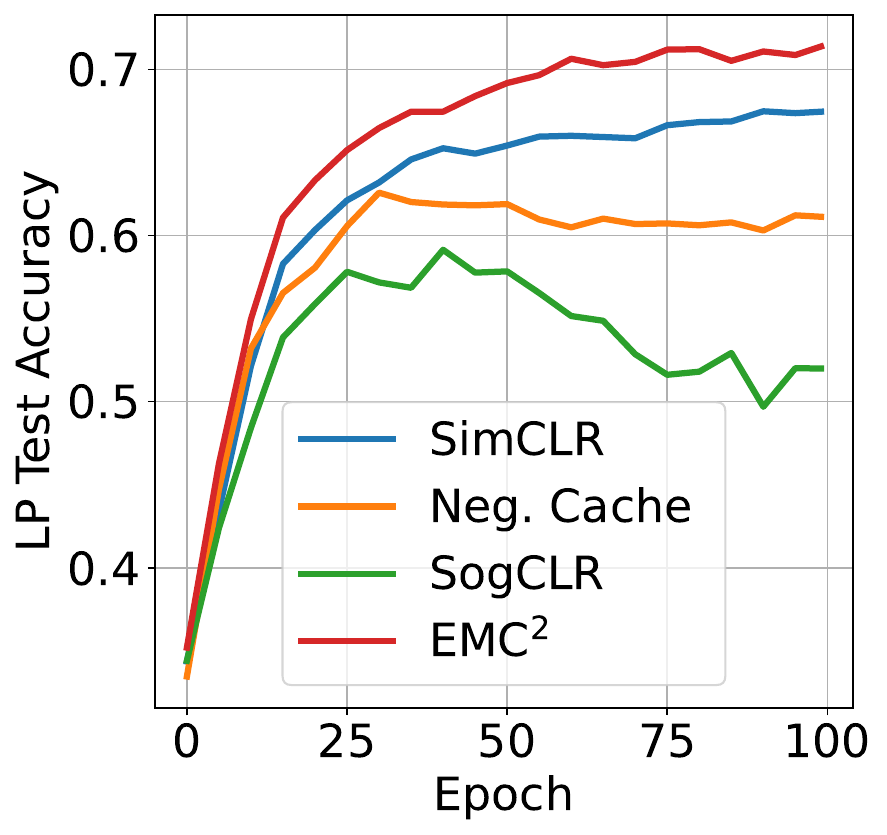}
    \includegraphics[width=0.238\textwidth]{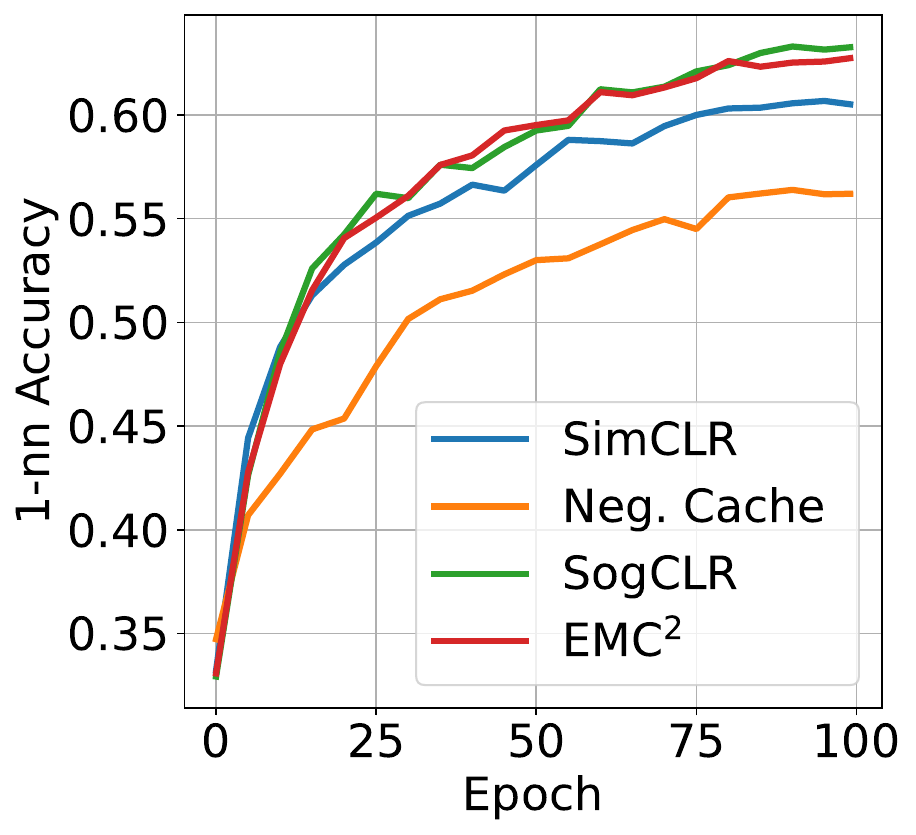}
    \\
    \includegraphics[width=0.238\textwidth]{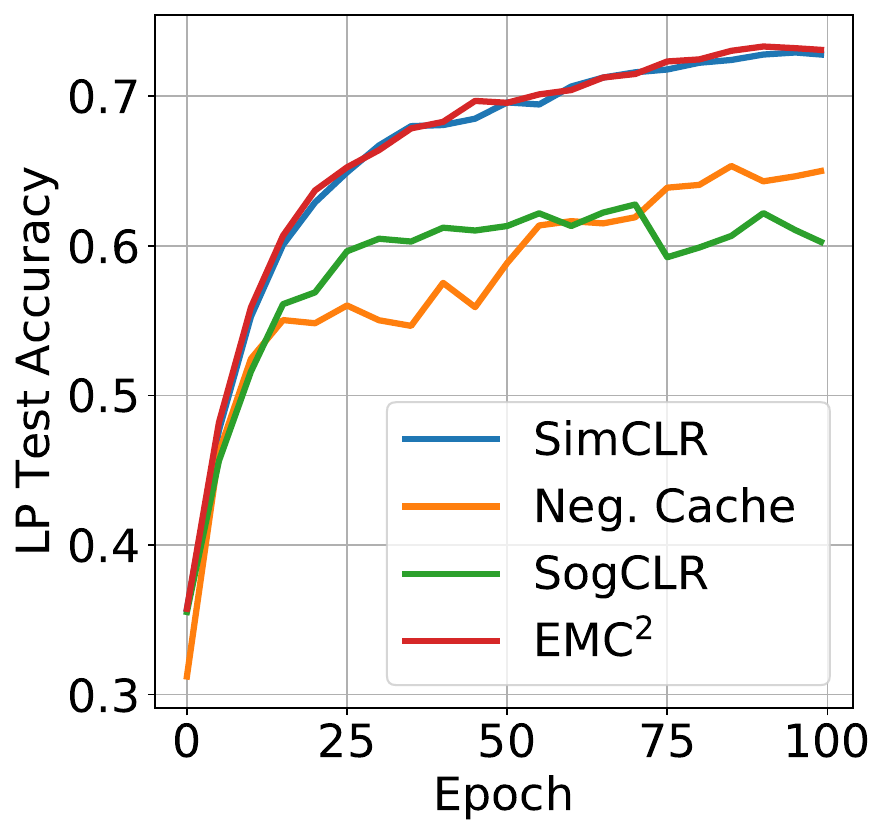}
    \includegraphics[width=0.238\textwidth]{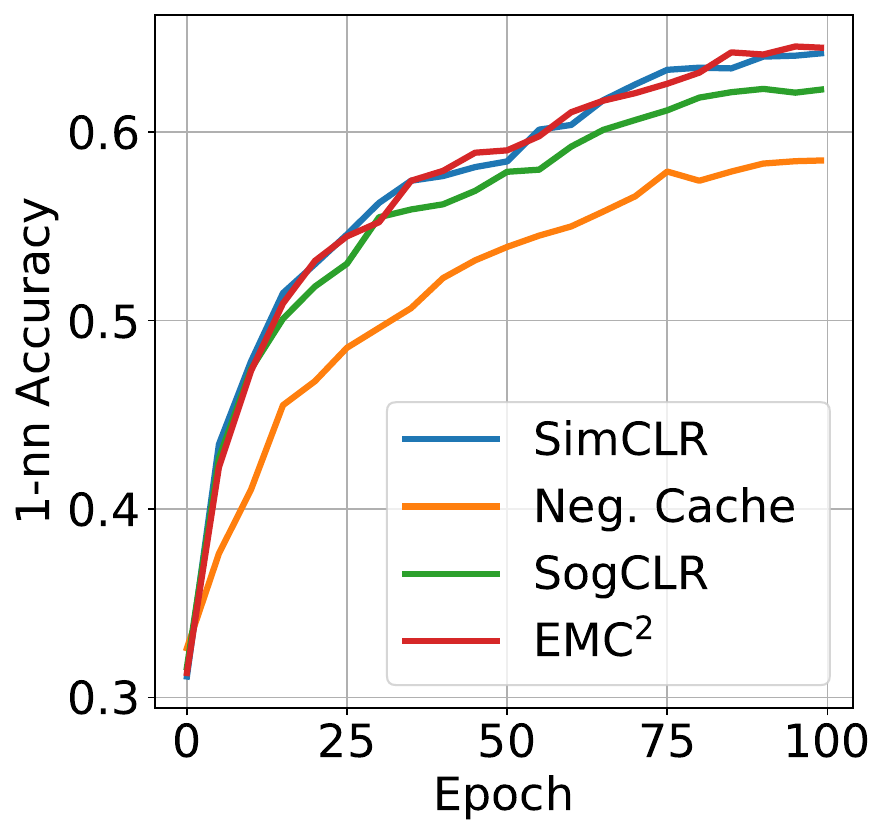}
    \caption{Comparison on {\tt STL-10} with ResNet-18 using batch size (top) $b = 32$, (bottom) $b = 256$. Horizontal axis is relative to the number of samples accessed.}\vspace{-.2cm}
    \label{fig:stl10}
\end{figure}

\subsection{Comparison to Baselines}
We compare {\algname} to baseline algorithms: Negative Cache \cite{lindgren2021efficient}, SimCLR \cite{chen2020simple}, SogCLR \cite{yuan2022provable}.
We first examine the effects of batch sizes $b \in \{ 32, 256 \}$ for the experiments on {\tt STL-10}, where a standard setting for this dataset is $b=256$, yet we note that a small batch size is often preferred for training large models due to limitations on GPU memory. Observe from Figure \ref{fig:stl10} that {\algname} consistently outperforms the other algorithms across all metrics regardless the choice of batch size. This is in line with the intuition that SimCLR is bottle-necked by a poor loss distribution on small batch size. For SogCLR, we suspect the performance drop on linear probe accuracy is due to the violation of feature homogeneity assumption on the setup of {\tt STL-10} with Resnet-18. For negative cache, its performance on using one negative with stale cache error does not keep up with in-batch negatives approaches.

\begin{figure}
    \centering \vspace{-0.15cm}
    \includegraphics[width=0.238\textwidth]{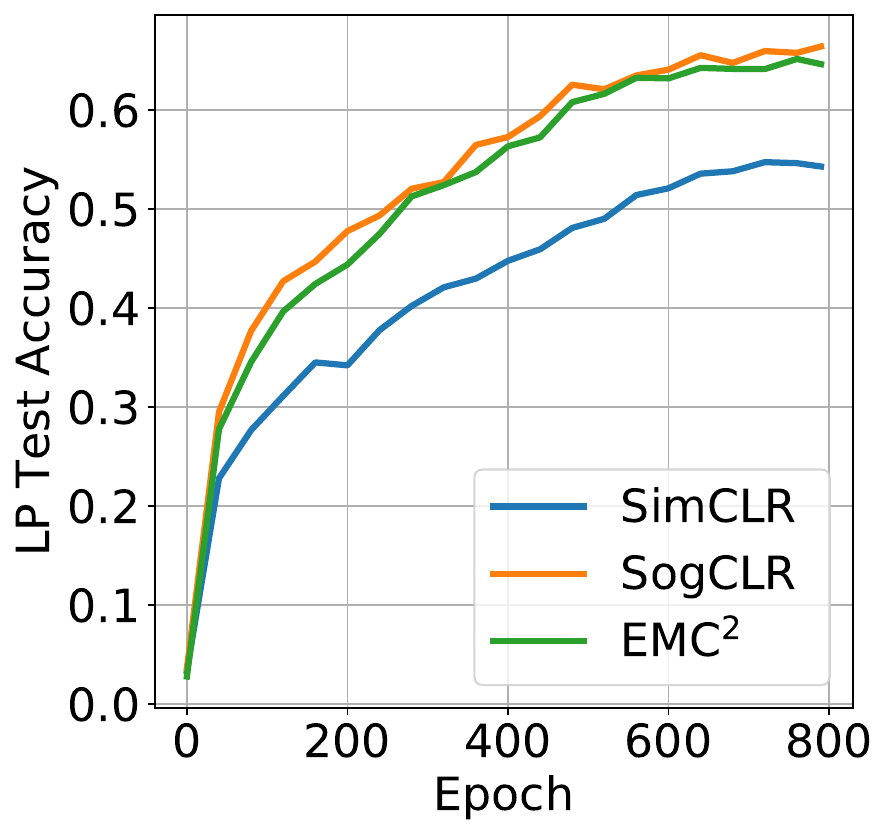}
    \includegraphics[width=0.238\textwidth]{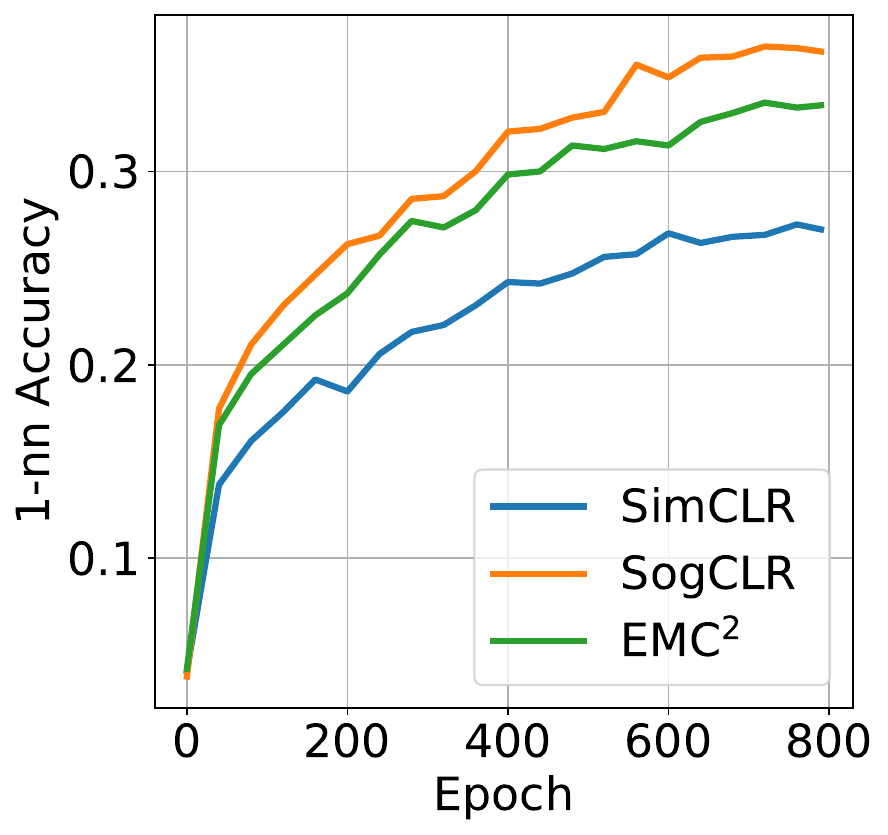}
    \caption{Comparison on {\tt Imagenet-100} with ResNet-50 using batch size $b=256$. Horizontal axis is relative to the number of samples accessed.}\label{fig:imagenet100}
\end{figure}

Figure~\ref{fig:imagenet100} shows a similar experiment but on the more difficult dataset {\tt Imagenet-100}. Note that the negative caching algorithm is not run due to the excess computation complexity in cache refreshing and Gumbel-max sampling. Observe that \algname~shows performance gain over SimCLR when trained on the standard batch size $b = 256$, while the performance is on par with SogCLR. 

From the above experiments, we observe that {\algname} delivers a consistent performance over different datasets, batch size, that is on par with the best compared baseline for the respective tasks. In Appendix \ref{app:time_plot}, additional time-complexity comparison on {\tt STL-10} is presented to demonstrate the practical benefits of \algname.

\begin{figure}
    \centering 
    \includegraphics[width=0.222\textwidth]{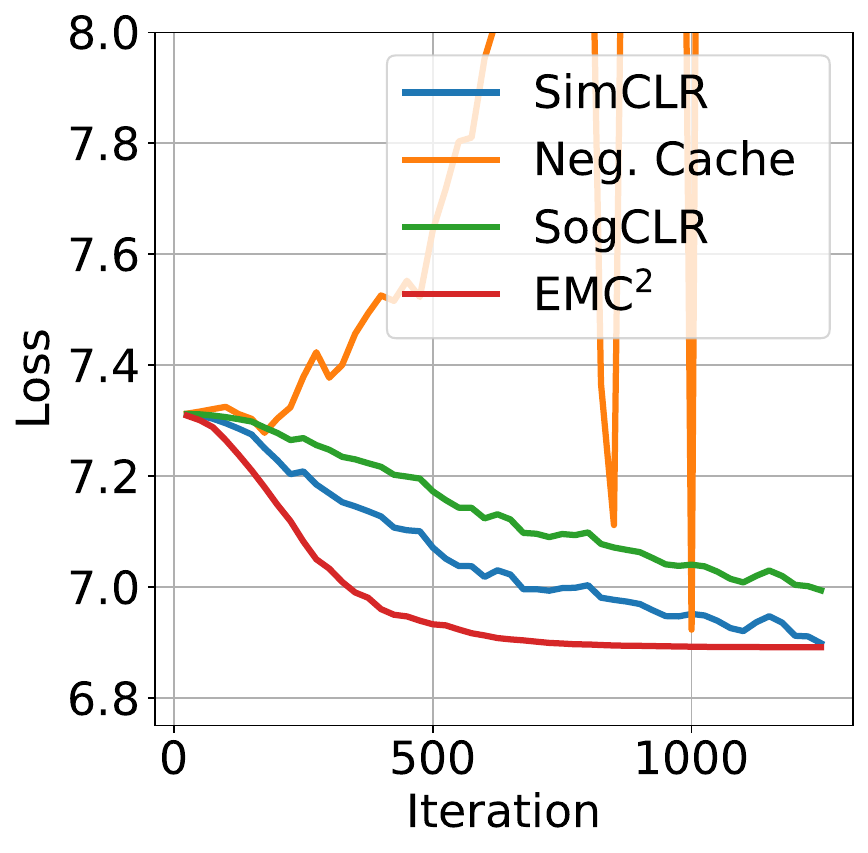}
    \includegraphics[width=0.238\textwidth]{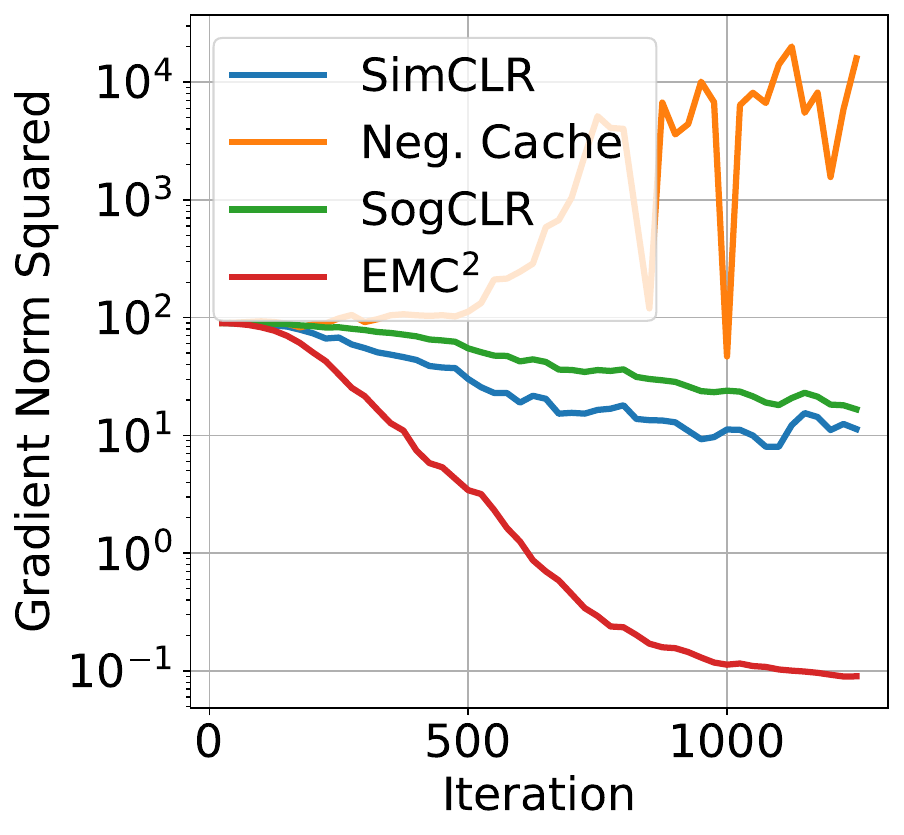}
    \caption{Comparison on a subset of {\tt STL-10} using the first 500 images and pre-computed two augmentations for each image. Trained using SGD with batch size $b = 4$.}
    \label{fig:stl10_subset_exp}
\end{figure}

\noindent \textbf{Convergence to Stationary Solution.}
To examine the accuracy of {\algname}, we construct a simple dataset (${\cal D}_{\rm pos}, {\bf D}_{\rm neg}$) by taking the first 500 images from {\tt STL-10} and using two pre-computed augmentations for each image. Under this setting, the exact value of ${\cal L}(\theta)$ in \eqref{eq:gcl} and $\| \nabla {\cal L}(\theta) \|^2 $ in \eqref{eq:gcl_grad_sum} can be evaluated as shown in Figure \ref{fig:stl10_subset_exp}. We observe that the solution found by {\algname} is more accurate than the other baselines by 2 orders of magnitude in terms of squared gradient norm.

\vspace{-.1cm}
\section{Conclusion}\vspace{-.1cm}
This paper proposed a novel method \algname~for optimizing the global contrastive loss \eqref{eq:gcl}. The algorithm combines an adaptively adjusted MCMC scheme for generating negative samples with a standard SGD update. We prove that {\algname} enjoys low memory and computation complexities, and admits a global convergence rate of ${\cal O}(1/\sqrt{T})$ towards a stationary solution for \eqref{eq:gcl}. Numerical experiments show that {\algname} enables small batch training for contrastive learning which is due to its global convergence property. 
We hope this work inspires future research into advanced sampling methods for contrastive learning, such as the Langevin dynamics, and the convergence property of optimization algorithms that rely on them.


\bibliography{reference}
\bibliographystyle{icml2024}

\newpage
\appendix
\onecolumn

\section{Proof of Lemma \ref{lemma:mcmc_rate}} \label{app:proof_mcmc_rate}
Notice that our target softmax distribution $\px$ has a finite support with each state having non-zero mass. For any $\theta \in \mathbb{R}^p$,
\begin{equation}
    \min_{x, z, Z} Q_{x,\theta}(z, Z) = \min_{x, z, Z} \frac{\exp(\beta ~ \phi(x;\theta)^\top \psi(z;\theta))}{\exp(\beta ~ \phi(x;\theta)^\top \psi(Z;\theta))} \stackrel{({\rm Assm.} ~\ref{assm:bounded_embd})}{\ge} \exp(-2c^2 \beta)
\end{equation}
Now recall that $\mneg = |\dneg(x)|$ for any $x \in \data_X$, i.e., the state space consists of $\mneg$ states on each Markov chain $Z_i$. Then the smallest probability to transition to state $z$ from any initial state is lower bounded by $\mneg^{-1}\exp(-2c^2 \beta)$, where the factor $\mneg^{-1}$ comes from using the uniform prior ${\rm Uniform}(\dneg(x))$ in Algorithm \ref{alg:multi_mcmc}. 

By Theorem 1.3 of \cite{mengersen1996rates}, the transition kernel of each Markov chain of $Z_i$ converges to its target distribution $\px$ at a geometric rate of $1 - \mneg^{-1}\exp(-2c^2 \beta)$. Therefore, the joint Markov chain of states $(Z_1, ..., Z_m)$ in Algorithm \ref{alg:multi_mcmc} has a mixing rate of $1 - m^{-1} \mneg^{-1}\exp(-2c^2 \beta)$ when the batch size $B=1$. As we perform mini-batch sampling of batch size $B$ on the expectation over $(x,y) \sim \dpos$ and $R$ steps of M-H algorithm over $z \sim \px$ \emph{in parallel}, the mixing rate can be improved to $[1 - B m^{-1} \mneg^{-1}\exp(-2c^2 \beta)]^R$. We can simplify the rate by the following upper bound, as
\begin{align}
    \left[1 - \frac{B}{ m \mneg \exp(2c^2 \beta) }\right]^R \leq \exp\left( - \frac{BR}{ m \mneg \exp(2c^2 \beta) }\right) \leq 1 - \frac{BR}{2 m \mneg \exp(2c^2 \beta)}  
\end{align}
where the first inequality uses $1-x \leq \exp(-x) ~\forall x \in \mathbb{R}$, the second inequality uses $\exp(-x) \leq 1 - x/2$ for $x \in [0, 1]$ and the assumption that $BR \leq m\mneg \exp(2c^2 \beta)$.
\hfill $\square$

\section{Proof of Lemma~\ref{lem:smooth}} \label{app:smooth}
For each Markov chain corresponding to $Z_i$ in the M-H algorithm, the acceptance probability is given by $\min\{1, Q_{x_i, \theta}( Z, Z' )\}$. To establish Lemma~\ref{lem:smooth}, we first observe that for any $Z,Z' \in \dneg(x_i)$ and $\theta, \theta' \in \mathbb{R}^p$, it holds
\begin{align}
    &|Q_{x,\theta}( Z, Z') - Q_{x,\theta'}( Z, Z')| \\
    &= |\exp\left(\beta\phi(x;\theta)^\top \psi(Z;\theta) - \beta\phi(x; \theta)^\top \psi(Z';\theta)\right) - \exp\left(\beta\phi(x;\theta')^\top \psi(Z;\theta') - \beta\phi(x; \theta')^\top \psi(Z';\theta')\right)| \\
    &\leq \exp(2c^2 \beta) \cdot \beta \cdot | \phi(x;\theta)^\top \psi(Z;\theta) - \phi(x; \theta)^\top \psi(Z';\theta) - (\phi(x;\theta')^\top \psi(Z;\theta') - \phi(x; \theta')^\top \psi(Z';\theta')) | \label{proof:exp_local_smooth} \\
    &\stackrel{({\rm Assm.}~ \ref{assm:smooth_kernel})}\leq 2 L_P \cdot \exp(2c^2 \beta) \cdot \beta \cdot \| \theta -\theta'\| \label{proof:lip_q}
\end{align}
where \eqref{proof:exp_local_smooth} uses the fact that $\exp(\cdot)$ is $\exp(2c^2 \beta)$-Lipschitz when restricted on the domain $[-2c^2 \beta, 2c^2 \beta]$ and the restriction is due to Assumption \ref{assm:bounded_embd}.

Now denote $\ker_{i,\theta} \in \mathbb{R}^{\mneg \times \mneg}$ as the transition matrix of $Z_i$ and
\begin{align}
    [\ker_{i,\theta}]_{z',z} = \begin{cases}
        \frac{1}{\mneg} \cdot \min \{1, Q_{x_i,\theta}(z,z') \} & \text{if}~ z \neq z', \\
        \frac{1}{\mneg} + \frac{1}{\mneg} \sum_{j \in [\mneg] \backslash \{z\}} (1-\min \{1, Q_{x_i,\theta}(z,j) \}) &\text{if}~ z = z'.
    \end{cases} \label{eq:kernel_value}
\end{align}
Then for $z \neq z'$,
\begin{align}
    |[\ker_{i,\theta}]_{z',z} - [\ker_{i,\theta'}]_{z',z}| &= \frac{1}{\mneg} |\min \{1, Q_{x_i,\theta}(z,z') \} - \min \{1, Q_{x_i,\theta'}(z,z') \}| \\
    &\leq \frac{1}{\mneg} |Q_{x_i,\theta}(z,z') -  Q_{x_i,\theta'}(z,z')| \stackrel{\eqref{proof:lip_q}}{\leq} \frac{1}{\mneg} \cdot 2 L_P \cdot \exp(2c^2 \beta) \cdot \beta \cdot \| \theta -\theta'\| 
\end{align}
and for $z = z'$,
\begin{align}
    |[\ker_{i,\theta}]_{z',z} - [\ker_{i,\theta'}]_{z',z}|
    &= \left| \frac{1}{\mneg} \sum_{j \in [\mneg] \backslash \{ z\} } ( - \min \{1, Q_{x_i,\theta}(z,j) \} + \min \{1, Q_{x_i,\theta'}(z,j) \} ) \right| \\
    &\leq \frac{1}{\mneg} \sum_{j \in [\mneg] \backslash \{ z\} } |Q_{x_i,\theta}(z,j) -  Q_{x_i,\theta'}(z,j)| \stackrel{\eqref{proof:lip_q}}{\leq} 2 L_P \cdot \exp(2c^2 \beta) \cdot \beta \cdot \| \theta -\theta'\| 
\end{align}

To quantify the smoothness of the transition kernel with $R$ Metropolis-Hastings steps, we observe that by the notation $[\ker]_{z,z'} = {\bf e}_z^\top \ker {\bf e}_{z'}$ for some basis vectors ${\bf e}_z, {\bf e}_{z'}$,
\begin{align}
    &\max_{z,z'} |{\tt P}_\theta(Z_i^{(0)}=z; Z_i^{(R)}=z') - {\tt P}_{\theta'}(Z_i^{(0)}=z; Z_i^{(R)}=z')| \\
    &= \max_{z,z'} |[\ker_{i,\theta}^R]_{z',z} - [\ker_{i,\theta'}^R]_{z',z}| \\
    &= \max_{z,z'} | {\bf e}_{z'}^\top( \ker_{i,\theta}^R  - \ker_{i,\theta'}\ker_{i,\theta}^{R-1} + \ker_{i,\theta'}\ker_{i,\theta}^{R-1} - \ker_{i,\theta'}^R ) {\bf e}_{z} | \\
    &\leq \max_{z,z'} | {\bf e}_{z'}^\top( \ker_{i,\theta}  - \ker_{i,\theta'})\ker_{i,\theta}^{R-1} {\bf e}_{z}| + | {\bf e}_{z'}^\top \ker_{i,\theta'} ( \ker_{i,\theta}^{R-1} - \ker_{i,\theta'}^{R-1} ){\bf e}_{z} | \\
    &\stackrel{(i)}{\leq} \max_{z,z'} \| {\bf e}_{z'}^\top( \ker_{i,\theta}  - \ker_{i,\theta'})\|_\infty \cdot \| \ker_{i,\theta}^{R-1} {\bf e}_{z} \|_1 + \| {\bf e}_{z'}^\top \ker_{i,\theta'} \|_1 \cdot \|  ( \ker_{i,\theta}^{R-1} - \ker_{i,\theta'}^{R-1} ){\bf e}_{z} \|_\infty  \\
    &\stackrel{(ii)}{\leq} \max_{z,z'} \| {\bf e}_{z'}^\top( \ker_{i,\theta}  - \ker_{i,\theta'})\|_\infty \stackrel{(iii)}{+} 2 \cdot |[\ker_{i,\theta}^{R-1}]_{z',z} - [\ker_{i,\theta'}^{R-1}]_{z',z}|  \\
    &\stackrel{\eqref{proof:lip_q}}{\leq} 2 L_P \cdot \exp(2c^2 \beta) \cdot \beta \cdot \| \theta -\theta'\| +  2 \cdot |[\ker_{i,\theta}^{R-1}]_{z',z} - [\ker_{i,\theta'}^{R-1}]_{z',z}| \\
    &\leq (1 + 2 + ... + 2^{R-1})  \cdot 2 L_P \cdot \exp(2c^2 \beta) \cdot \beta \cdot \| \theta -\theta'\| \\
    & = (2^R - 1) \cdot 2 L_P \cdot \exp(2c^2 \beta) \cdot \beta \cdot \| \theta -\theta'\| \label{proof:smooth_r_step_kernel}
\end{align}
where $(i)$ uses Hölder's inequality for the norm pair $(\| \cdot \|_1, \| \cdot \|_\infty)$, $(ii)$ uses the fact that $\ker_{i,\theta}^{R-1}$ is a column stochastic matrix and $(iii)$ uses the inequality $\| {\bf e}_{z'}^\top \ker_{i,\theta'} \|_1 \leq 2$ because
\begin{align}
    &\| {\bf e}_{z'}^\top \ker_{i,\theta'} \|_1 = \sum_{z \in [\mneg]} [\ker_{i,\theta}]_{z',z}\\
    &\stackrel{\eqref{eq:kernel_value}}{=} \sum_{z\in[\mneg] \backslash \{z'\}} \frac{1}{\mneg} \min\{ 1, Q_{x_i,\theta}(z,z')\} \label{eq:lose_bound} \\
    &\quad + \frac{1}{\mneg} +  \frac{1}{\mneg} \sum_{j \in [\mneg] \backslash \{z'\}} (1-\min \{1, Q_{x_i,\theta}(z',j) \}) \\
    &\leq \frac{\mneg - 1}{\mneg} + \frac{1}{\mneg} + \frac{\mneg-1}{\mneg} \leq  2
\end{align}

Finally, in the Markov chain $\tilde{\bm{\xi}}_0 \to \tilde{\bm{\xi}}_1 \to \cdots$, since only $B$ state variables $\{ Z_{i_k^{(t)}} \}_{k=1}^B$ are active or updated in each transition, we have 
\begin{equation} 
    {\tt P}_{\theta} ( \bm{\xi} ; \bm{\xi}' ) = 
    \begin{cases}
    \quad 0 & \text{if} ~\Delta_\xi = \| \xi - \xi' \|_0 > B, \\
        \left(\begin{matrix}m \\ B\end{matrix}\right)^{-1} \cdot \left(\begin{matrix}m - \Delta_\xi \\ B - \Delta_\xi \end{matrix}\right) \cdot \prod_{k=1}^B {\tt P}_\theta (Z_{i_k^{(t)}}^{(0)}; Z_{i_k^{(t)}}^{(R)}) & \text{otherwise}.
    \end{cases}
\end{equation}
where $\left(\begin{matrix}m \\ B\end{matrix}\right)^{-1} \left(\begin{matrix}m - \Delta_\xi \\ B - \Delta_\xi \end{matrix}\right)$ is the probability that $j \in \{ i_k^{(t)} \}_{k=1}^B$ for $j$ s.t. $\xi(j) \neq \xi'(j)$. To simplify notations, suppose $Z_{i_k^{(t)}}^{(0)} = z_{t,k}$ and $Z_{i_k^{(t)}}^{(R)} = z_{t,k}'$, then
\begin{align}
    &|{\tt P}_{\theta} ( \xi ; \xi' ) - {\tt P}_{\theta'} ( \xi ; \xi' )| = \left(\begin{matrix}m \\ B\end{matrix}\right)^{-1} \left(\begin{matrix}m - \Delta_\xi \\ B - \Delta_\xi \end{matrix}\right) \left|\prod_{k=1}^B {\tt P}_\theta (Z_{i_k^{(t)}}^{(0)}; Z_{i_k^{(t)}}^{(R)}) - \prod_{k=1}^B {\tt P}_{\theta'} (Z_{i_k^{(t)}}^{(0)}; Z_{i_k^{(t)}}^{(R)}) \right| \\
    & = \left(\begin{matrix}m \\ B\end{matrix}\right)^{-1} \left(\begin{matrix}m - \Delta_\xi \\ B - \Delta_\xi \end{matrix}\right) \left|\prod_{k=1}^B  [\ker_{i,\theta}^R]_{z_{t,k}, z_{t,k}'} - \prod_{k=1}^B [\ker_{i,\theta'}^R]_{z_{t,k}, z_{t,k}'} \right| \\
    &= \left(\begin{matrix}m \\ B\end{matrix}\right)^{-1} \left(\begin{matrix}m - \Delta_\xi \\ B - \Delta_\xi \end{matrix}\right) \Big| \prod_{k=1}^B [\ker_{i,\theta}^R]_{z_{t,k}, z_{t,k}'}- [\ker_{i,\theta'}^R]_{z_{t,1}, z_{t,1}'} \prod_{k=2}^B [\ker_{i,\theta}^R]_{z_{t,k}, z_{t,k}'} \\
    &\quad + [\ker_{i,\theta'}^R]_{z_{t,1}, z_{t,1}'} \prod_{k=2}^B [\ker_{i,\theta}^R]_{z_{t,k}, z_{t,k}'} - \prod_{k=1}^B [\ker_{i,\theta'}^R]_{z_{t,k}, z_{t,k}'} \Big| \\
    &\stackrel{\eqref{proof:smooth_r_step_kernel}}{\leq} \left(\begin{matrix}m \\ B\end{matrix}\right)^{-1} \left(\begin{matrix}m - \Delta_\xi \\ B - \Delta_\xi \end{matrix}\right)  B \cdot 2^{R+1} \cdot L_P \cdot \exp(2c^2 \beta) \cdot \beta \cdot \| \theta -\theta'\|
\end{align}
The proof is concluded by taking maximum over $0 \leq \Delta_\xi \leq B$. \hfill $\square$

\section{Proof of Theorem \ref{thm:main}} \label{app:proof_thm}
Our idea is to apply Theorem~2 of \cite{karimi2019non} which shows the convergence of biased stochastic approximation scheme with Markovian noise such as \eqref{eq:sgd}. In particular, we shall take the Lyapunov function therein as our global contrastive loss, i.e., $V(\theta) = {\cal L}(\theta)$. We proceed by verifying the required assumptions. 

\paragraph{Verifying A1-A2 of \cite{karimi2019non}} Using Lemma~\ref{lemma:mcmc_rate}, we observe that the mean field of the stochastic update in \eqref{eq:sgd}, i.e., the expected value of ${\cal H}( \xi; \theta )$ when $\xi$ is drawn from the stationary distribution of the Markov chain induced by ${\tt P}_{\theta}$, coincides with the gradient of ${\cal L}(\theta)$, i.e., $h(\theta) = \nabla {\cal L}(\theta)$. As such, A1-A2 of \cite{karimi2019non} are satisfied with $c_0 = d_0 = 0$, $c_1 = d_1 = 1$.

\paragraph{Verifying A3 of \cite{karimi2019non}} For this, we need to show that $\nabla {\cal L}( \theta)$ is Lipschitz continuous. We observe that a stronger condition holds as the stochastic gradient map is Lipschitz w.r.t.~$\theta$. For any sample $x, y, z$ and any $\theta, \theta' \in \mathbb{R}^p$, we have
\begin{align}
    &\| - \beta \nabla_\theta (\phi(x;\theta)^\top \psi(y;\theta) ) + \beta \nabla_\theta (\phi(x;\theta)^\top \psi(z;\theta)) - [- \beta \nabla_\theta (\phi(x;\theta')^\top \psi(y;\theta') ) + \beta \nabla_\theta (\phi(x;\theta')^\top \psi(z;\theta'))] \| \\
    &\leq \beta \| \nabla_\theta (\phi(x;\theta)^\top \psi(y;\theta) ) - \nabla_\theta (\phi(x;\theta')^\top \psi(y;\theta') ) \| + \beta \|\nabla_\theta (\phi(x;\theta)^\top \psi(z;\theta)) - \nabla_\theta (\phi(x;\theta')^\top \psi(z;\theta')) \| \notag \\
    &\stackrel{({\rm Assm.}~\ref{assm:smooth_sgrad})}{\leq} 2 \beta L_H \|\theta - \theta'\| \notag
\end{align}

\paragraph{Verifying A5-6 of \cite{karimi2019non}} For these assumptions, we observe that A12-A14 of \cite{karimi2019non} can be satisfied with the constants
\begin{equation}
\bar{L}_P = 2^{R+1} B L_P \cdot \exp( 2 c^2 \beta ) \cdot \beta, \quad \bar{L}_H = 2 \beta L_H, \quad \bar{\rho} = 1 - \frac{BR}{2 m \mneg \exp(2c^2 \beta)} , \quad K_P = 1, \quad \bar{\sigma} \stackrel{\eqref{proof:sigma_bar}}{=}  2\beta \sigma
\end{equation}
Applying Lemma~7 of \cite{karimi2019non} shows that A5-A6 can be satisfied with 
\begin{equation}
    L_{PH}^{(0)} = \frac{2 \beta \sigma \bar{\rho}}{ BR m^{-1} \mneg^{-1} \exp(-2c^2 \beta) / 2 }, \quad L_{PH}^{(1)} = \frac{6 \cdot 2^{R+1} B \exp(2c^2 \beta) \beta^2 \sigma L_P}{( BR m^{-1} \mneg^{-1} \exp(-2c^2 \beta) / 2 )^2} +  \frac{2 \beta L_H}{ BR m^{-1} \mneg^{-1} \exp(-2c^2 \beta) / 2 }
\end{equation}

\paragraph{Verifying A7 of \cite{karimi2019non}} The stochastic gradient used in \eqref{eq:sgd} has a uniformly bounded error from its mean field as
\begin{align}
    &\left\| {\cal H}( \xi_{t+1} ; \theta_t) - \nabla {\cal L}(\theta_t) \right\| \\
    &\leq \beta \left\| \frac{1}{B} \sum_{\ell=1}^B H(x^{(t)}_{i_{\ell}}, y^{(t)}_{i_{\ell}}; \theta_t) - \mathbb{E}_{(x,y) \sim \dpos }[\nabla_\theta(\phi(x;\theta)^\top \psi(y;\theta)) ] \right\| \notag \\
    &\quad + \beta \left\| \frac{1}{B (B-P) }\sum_{k=1}^B \sum_{r=P}^{R-1} H( x_{i_k^{(t)}}, \tilde{Z}_{i_k^{(t)}}^{(r)} ; \theta_t ) - \mathbb{E}_{(x,y) \sim \dpos}  \mathbb{E}_{z \sim \px} [\nabla_\theta(\phi(x;\theta)^\top \psi(z;\theta)) ] \right\| \notag \\
    &\leq \frac{\beta}{B} \sum_{\ell=1}^B \|  H(x^{(t)}_{i_{\ell}}, y^{(t)}_{i_{\ell}}; \theta_t) - \mathbb{E}_{(x,y) \sim \dpos }[\nabla_\theta(\phi(x;\theta)^\top \psi(y;\theta)) ] \| \notag \\
    &\quad + \frac{\beta}{B (B-P) }\sum_{k=1}^B \sum_{r=P}^{B-1}  \| H( x_{i_k^{(t)}}, \tilde{Z}_{i_k^{(t)}}^{(r)} ; \theta_t ) - \mathbb{E}_{(x,y) \sim \dpos}  \mathbb{E}_{z \sim \px} [\nabla_\theta(\phi(x;\theta)^\top \psi(z;\theta)) ] \| \notag \\
    &\stackrel{({\rm Assm. }~ \ref{assm:bounded_var})}{\leq} 2 \beta \sigma \label{proof:sigma_bar}
\end{align}

\paragraph{Convergence of \eqref{eq:sgd}} Upon verifying A1-A3, A5-A7 of \cite{karimi2019non}, we can apply Theorem 2 therein to analyze the convergence of \eqref{eq:sgd}. In particular, we choose a constant step size $\gamma$ satisfying $\gamma \leq \frac{1}{2(2\beta L_H + C_h)}$. For any $T \geq 1$ and the randomly drawn $t \sim {\rm Uniform}([0, ..., T-1])$, we have
\begin{align}
\mathbb{E}[\| \nabla {\cal L}(\theta_t) \|^2] & = \frac{1}{T} \sum_{t=0}^{T-1} \mathbb{E}[\| \nabla {\cal L}(\theta_t) \|^2] \\
& \leq  \frac{\mathbb{E}[{\cal L}(\theta_0) - {\cal L}(\theta_T)] + 3 L_{PH}^{(0)} + [8 \beta^3 L_H \sigma^2 + 2 L_{PH}^{(1)}\beta \sigma + 2 \beta L_H L_{PH}^{(0)} (1 + 2\beta \sigma)](\gamma^2 T)}{\gamma T / 2}
\end{align}
where $C_h = L_{PH}^{(1)} (1 + \sigma) + L_{PH}^{(0)}(\beta L_H + 1)$.
Simplifying and rearranging terms leads to the conclusion. 

\paragraph{Estimating $\mathbb{E}[{\cal L}(\theta_0) - {\cal L}(\theta_T)]$} We observe that by Assumption \ref{assm:bounded_embd}, for any $\theta \in \mathbb{R}^p$,
\begin{align}
    {\cal L}(\theta) &= \mathop{\mathbb{E}}_{(x,y) \sim \dpos }\left[-\beta ~\phi(x;\theta)^\top \psi(y;\theta) \right] + \mathop{\mathbb{E}}_{(x,y) \sim \dpos }\left[ \log \sum_{z \in \dneg(x)} \exp(\beta ~ \phi(x;\theta)^\top \psi(z;\theta)) \right] \\
    &\leq \beta + \mathop{\mathbb{E}}_{(x,y) \sim \dpos }\left[ \log \sum_{z \in \dneg(x)} \exp(\beta) \right] \notag
\end{align}
By a similar argument,
\begin{equation}
    {\cal L}(\theta) \ge -\beta + \mathop{\mathbb{E}}_{(x,y) \sim \dpos }\left[ \log \sum_{z \in \dneg(x)} \exp(-\beta) \right]
\end{equation}
Therefore,
\begin{align}
\mathbb{E}[{\cal L}(\theta_0) - {\cal L}(\theta_T)] &\leq 2 \beta + \mathop{\mathbb{E}}_{(x,y) \sim \dpos }\left[\log \sum_{z \in \dneg(x)} \exp(\beta) - \log \sum_{z \in \dneg(x)} \exp(-\beta) \right]\\
&= 2 \beta + \mathop{\mathbb{E}}_{(x,y) \sim \dpos }\left[\log \left( \frac{\sum_{z \in \dneg(x)} \exp(\beta)}{ \sum_{z \in \dneg(x)}  \exp(-\beta)} \right) \right] \notag \\
&= 2 \beta + \mathop{\mathbb{E}}_{(x,y) \sim \dpos }\left[\log \exp(2\beta) \right] = 4\beta. \notag
\end{align}
\hfill $\square$

\section{Experiment Details} \label{app:exp}
\subsection{Details of Dataset and Hyperparameters}
In Table \ref{table:dataset}, we list the important attributes of datasets we used in the experiment section.
\begin{table}[h] \centering
\begin{tabular}{l|l|l|l}\toprule
{\bf Dataset} & {\bf \# of Pos. Pairs $m$}& {\bf \# of Neg. Samples $\mneg$} & {\bf Image Crop Size} 
\\ \midrule 
{\tt STL-10}       & 100,000 (inf. aug) & 100,000 (inf. aug) & 96 $\times$ 96 \\
{\tt Imagenet-100} & 124,689 (inf. aug) & 124,689 (inf. aug) & 224 $\times$ 224 \\
\bottomrule
\end{tabular}
\caption{\label{table:dataset} Datasets attributes.}
\end{table}

In Table \ref{table:hyperprms}, we list the hyperparameter values adopted in our experiments.

\begin{table}[h] 
\centering
\resizebox{.99\linewidth}{!}{\begin{tabular}{l|l|l|l|l|l|l|l|l}\toprule
{\bf Dataset} & {\bf Model} & \tworows{\bf Inverse}{{\bf Temp.} $\beta$}& \tworows{\bf Batch}{{\bf Size} $b$} & \tworows{\bf Learning}{{\bf Rate} $\gamma$} &  \tworows{\bf Feature }{{\bf Dim.} $d$}  &\tworows{\bf Weight}{\bf Decay} & \tworows{{\bf Cache Refresh} $\rho$}{ \bf (Negative Cache)} & \tworows{{\bf Burn-in Steps} $P$}{\bf (\algname)} \\ \midrule 
{\tt STL-10}       & Resnet-18 & 14.28 & 32     & $10^{-4}$                & 512                         & $10^{-4}$  & 0.01024  & 31  \\
{\tt STL-10}       & Resnet-18 & 5 &256     & $10^{-3}$                & 512                         & $10^{-4}$  & 0.1 & 255   \\
{\tt Imagenet-100} & Resnet-50 & 14.28 &256     & 1.2                  & 128                         & $10^{-6}$ &   -   & 255  \\
{\tt STL-10} subset    & Resnet-18 & 5 &4     & $10^{-3}$                & 512                         & $10^{-4}$  & 0.1 & 3   \\
\bottomrule
\end{tabular}}

\caption{\label{table:hyperprms} Hyperparameter values of the experiments.}
\end{table}

\subsection{Run-Time Complexity} \label{app:time_plot}
\begin{figure}[H]
    \centering \vspace{-0.15cm}
    \includegraphics[width=0.245\textwidth]{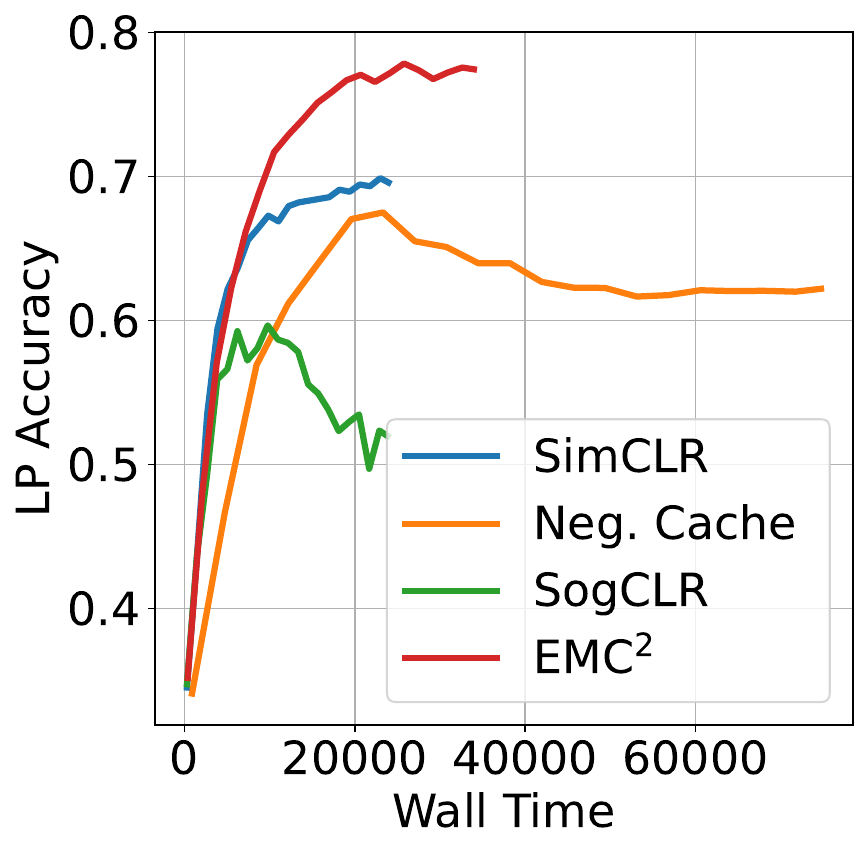}
    \includegraphics[width=0.245\textwidth]{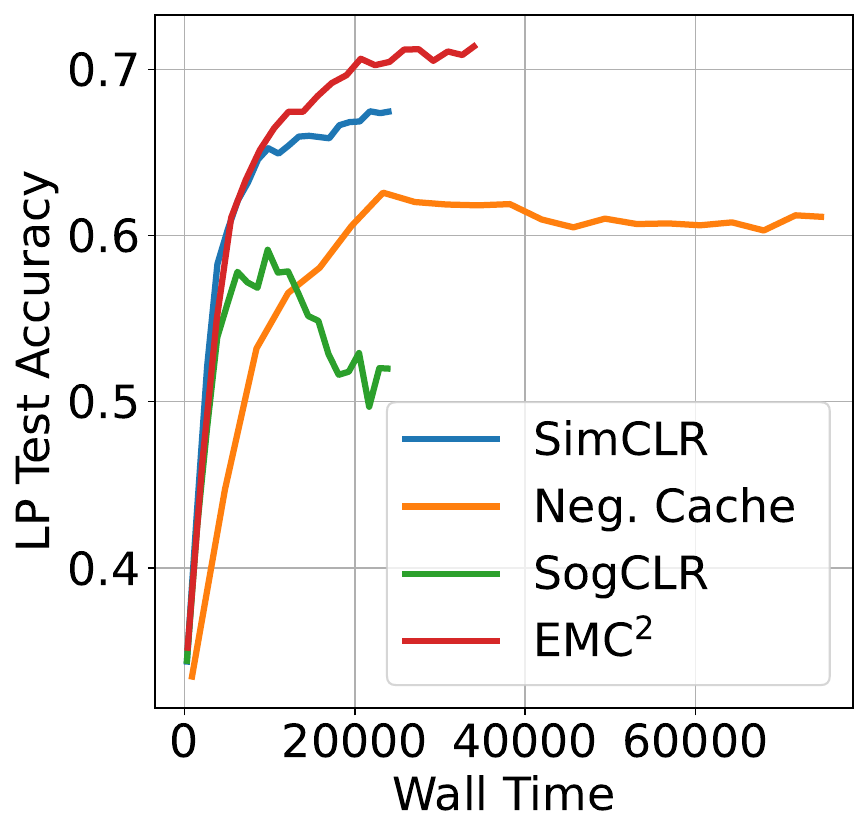}
    \includegraphics[width=0.245\textwidth]{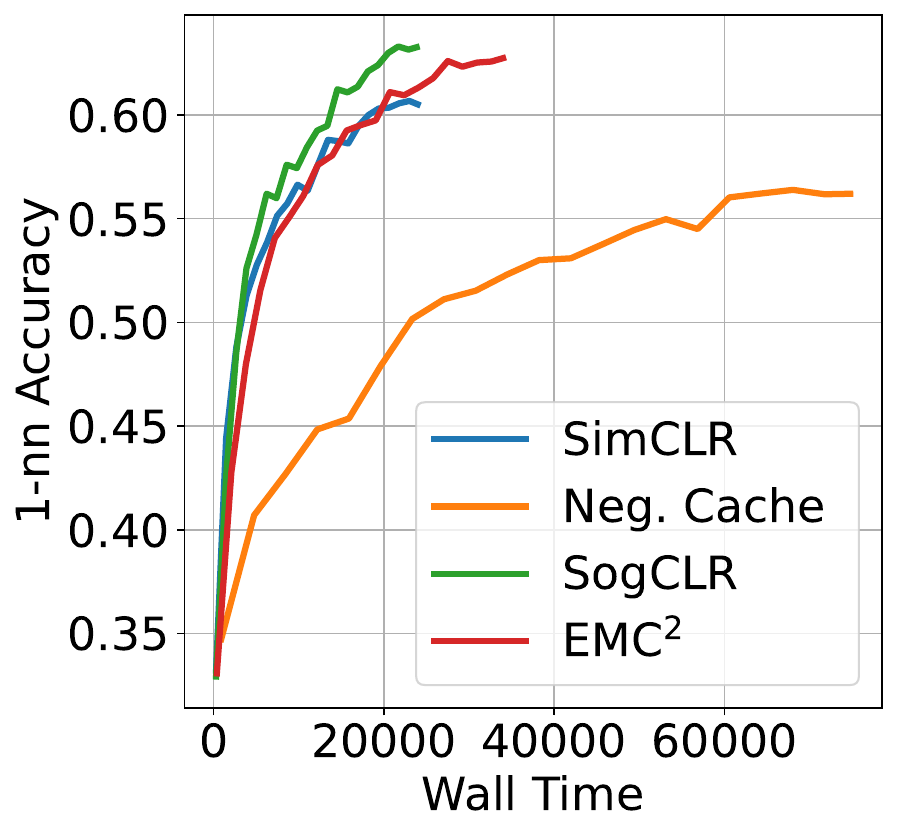}
    \includegraphics[width=0.245\textwidth]{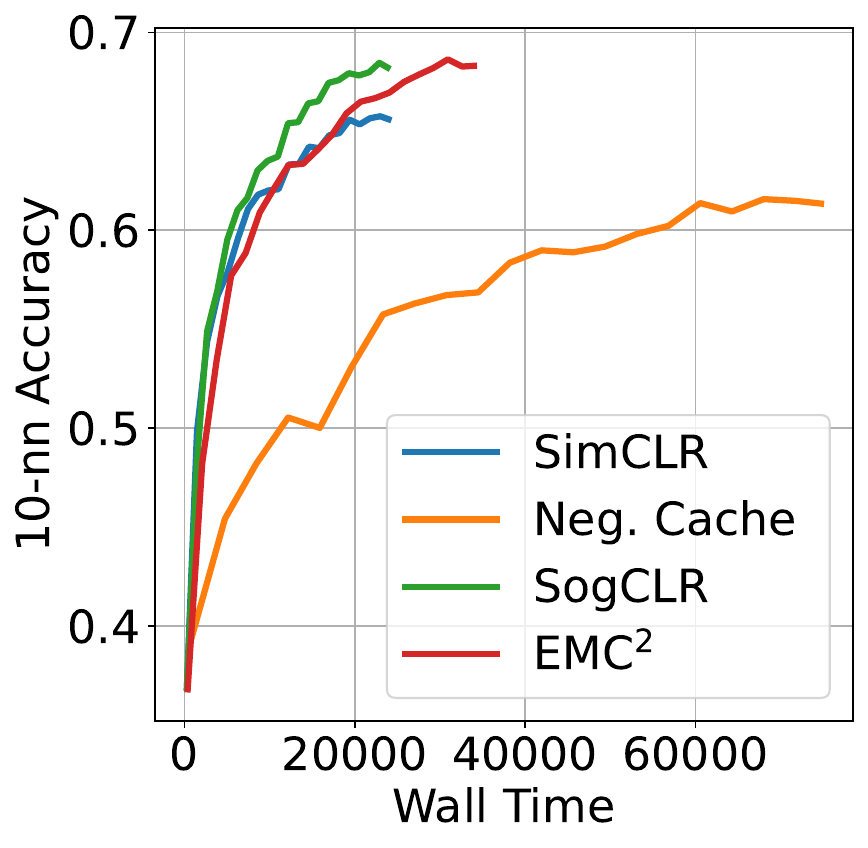}
    \caption{Training 100 epochs on {\tt STL-10} with ResNet-18 using batch size $b = 32$. Horizontal axis is relative to the wall-clock training time in seconds.}
    \label{fig:stl10_time}
\end{figure}

Since the computational complexity differs among algorithms, we provide a performance-time plot in Figure \ref{fig:stl10_time} to compare the time complexity. Note that in this setup, negative cache algorithm uses four Tesla T4 GPUs for training and refreshing the negative cache while the other algorithms run on one Tesla T4 GPU.

{~}


\end{document}